%% file: example_paper.tex
\documentclass{article}

\usepackage{microtype}
\usepackage{graphicx}
\usepackage{subfigure}
\usepackage{booktabs} % for professional tables

\usepackage{hyperref}

\usepackage[accepted]{icml2023}

\usepackage{amsmath}
\usepackage{amssymb}
\usepackage{mathtools}
\usepackage{amsthm}
\usepackage{paralist}

\usepackage{xcolor}

\usepackage[capitalize,noabbrev]{cleveref}

\input{defs}
\input{math_commands.tex}

\usepackage{url}            % simple URL typesetting
\DeclareUrlCommand\url{\color{blue!50}} % use blue url color

\theoremstyle{plain}

\theoremstyle{definition}
\theoremstyle{remark}
\usepackage[textsize=tiny]{todonotes}

\icmltitlerunning{Taxonomy-Structured Domain Adaptation}

\begin{document}

\twocolumn[
\icmltitle{Taxonomy-Structured Domain Adaptation}



\icmlsetsymbol{equal}{*}

\begin{icmlauthorlist}
\icmlauthor{Tianyi Liu}{equal,rut}
\icmlauthor{Zihao Xu}{equal,rut}
\icmlauthor{Hao He}{mit}
\icmlauthor{Guang-Yuan Hao}{cuhk}
\icmlauthor{Guang-He Lee}{}
\icmlauthor{Hao Wang}{rut}
% \icmlauthor{Firstname7 Lastname7}{mit}
% \icmlauthor{Firstname8 Lastname8}{rut,mit}
\end{icmlauthorlist}

\icmlaffiliation{rut}{Rutgers University}
\icmlaffiliation{mit}{Massachusetts Institute of Technology}
\icmlaffiliation{cuhk}{The Chinese University of Hong Kong}

\icmlcorrespondingauthor{Tianyi Liu}{tl579@scarletmail.rutgers.edu}
\icmlcorrespondingauthor{Zihao Xu}{zihao.xu@rutgers.edu}
\icmlcorrespondingauthor{Hao Wang}{hw488@cs.rutgers.edu}

\icmlkeywords{Machine Learning, ICML}

\vskip 0.3in
]



\printAffiliationsAndNotice{\icmlEqualContribution} % otherwise use the standard text.

\begin{abstract}
Domain adaptation aims to mitigate distribution shifts among different domains. However, traditional formulations are mostly limited to categorical domains, greatly simplifying nuanced domain relationships in the real world. In this work, we tackle a generalization with taxonomy-structured domains, which formalizes domains with nested, hierarchical similarity structures such as animal species and product catalogs. We build on the classic adversarial framework and introduce a novel \emph{taxonomist}, which competes with the adversarial discriminator to preserve the taxonomy information. The equilibrium recovers the classic adversarial domain adaptation's solution if given a non-informative domain taxonomy (e.g., a flat taxonomy where all leaf nodes connect to the root node) while yielding non-trivial results with other taxonomies. Empirically, our method achieves state-of-the-art performance on both synthetic and real-world datasets with successful adaptation. Code is available at \url{https://github.com/Wang-ML-Lab/TSDA}.
\end{abstract}

\section{Introduction} \label{sec:intro}
% When applying machine learning models to real world problem, we usually neglect the inherent assumption that training data and testing data shares identical distribution.
% Generalizing 
% To ensure the generalization ability of 
% \ghl{new below HW: finished a pass}

Learning generalizable models is a central goal in machine learning. The majority of the literature has been devoted to the standard i.i.d. setting where training data and testing data are assumed to have the same distribution. However, many real world problems inherently exhibit distributional shifts. For example, to make commercial impacts, a company must be able to reach brand-new groups of users~\cite{ZESRec}, whose data are not likely to be the same as the data used to fit production models. As a rapidly emerging sub-field, transfer learning aims to tackle this problem~\cite{pan2010survey,transfer_learning}. 

A series of work has attempted to accomplish transfer learning via domain adaptation~\cite{TLSurvey,MMD,CDAN,MCD,GTA,MDD,moment,blending,domain_bridge,DA_distillation}, i.e., learning representations whose distribution is well aligned across different domains. This is typically done through learning against an adversary who tries to distinguish different domains. While the approach is widely studied both theoretically and empirically~\cite{redko2020survey,ben2010theory,AMSDA,MDD,InvariantDA}, the representation of domains is almost entirely limited to simple categories. This is problematic since categorical variables do not admit any meaningful measurements of similarity or distance. Indeed, an ideal transfer learning approach should be able to control the transferability depending on the similarity across domains. For instance, if we treat dog breeds as domains, an ideal video segmentation system for bassets should behave more similarly for beagles than for pomeranians. While existing adversarial approach mitigates the shifts among domains, the solution can still be inadequate since it ignores the potential structure among domains. 

To this end, we extend domain adaptation to taxonomy-structured domains (see~\figref{fig:im} and \figref{fig:cub} for some example taxonomies). 
% We provide an example taxonomy in \blue{Figure ?} for illustration. 
Taxonomies abound in our culture for categorizing items, ranging from biological studies to library classification systems. Mathematically, ``taxonomy'' is a \emph{nested} hierarchical representation~\cite{rangapuram2023coherent}. Each \emph{node} in the taxonomy specifies a level of \emph{invariance} that holds for the domains within the node, which can be further broken down to a lower level of invariance in its child nodes. Finally, the lowest level of invariance is simply a single domain. We emphasize that a taxonomy of domains is different from a tree of domains, as a non-leaf node in a taxonomy typically involves multiple domains. The specification of invariance among domains thus becomes nested, which can be naturally translated to the induced similarities. Indeed, the similarity between two domains can be easily gauged through their common ancestors, which are also nested. 

In this work, we propose Taxonomy-Structured Domain Adaptation (TSDA), a plug-in extension of adversarial domain adaption by introducing a novel \emph{taxonomist}, who guides the representation to exhibit the given domain taxonomy. This enables representation learning with a flexible balance between domain similarity and domain invariance. Despite the obvious contradictory nature, it is evident that either extreme fails to capture some important inductive biases in learning. Indeed, absolute invariance to the domain prevents the model from leveraging similarities among domains to improve statistical efficiency, while purely relying on domain similarities can easily pick up spurious correlations among different domains. The flexibility allows us to achieve a suitable trade off according to the predictive task. We summarize our contributions are as follows:

% \begin{compactitem}
\begin{compactitem}
\item We identify the problem of adaptation across taxonomy-structured domains and develop taxonomy-structured domain adaptation (TSDA) as the first general DA method to address this problem. 
\item Our theoretical analysis shows that a natural extension of typical DA methods fails to take advantage of domain similarity reflected in a domain taxonomy and degenerates to uniform alignment. 
\item We further prove that TSDA retains typical DA methods' capability of uniform alignment when the domain taxonomy is non-informative, and balances domain similarity and domain invariance for other taxonomies. 
\item Empirical results show that our TSDA improves upon state-of-the-art DA methods on both synthetic and real-world datasets. 
\end{compactitem}

\section{Related Work}
\begin{figure}
\centering
\vskip -0.00in
% \subfigure{
\includegraphics[width=\textwidth,height=3.1cm,keepaspectratio]{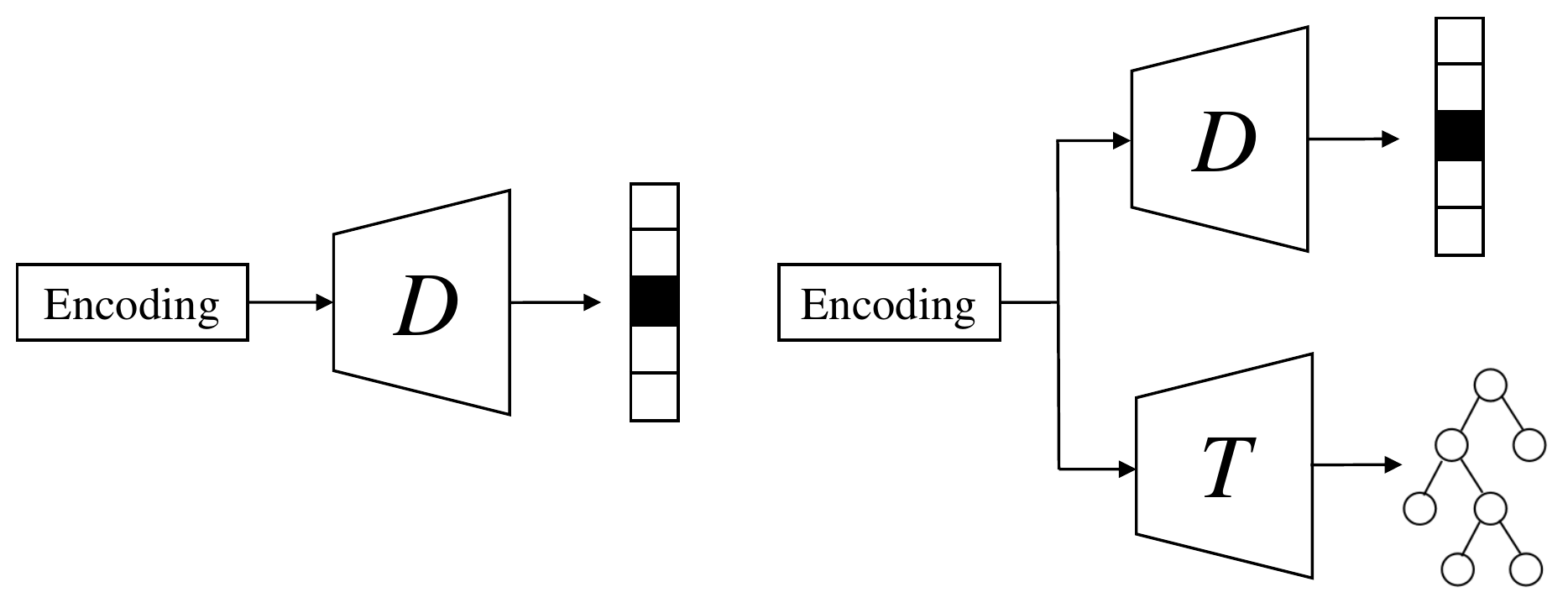}
% }
% \subfigure{
% \includegraphics[width=0.55\textwidth]{fig/StruDA-H-Indoor.pdf}}
\vskip -0.1cm% -0.2cm
% \captionsetup{font={scriptsize}}
\caption{Difference between typical DA methods and TSDA. \textbf{Left:} In traditional DA methods, the discriminator classifies the domain index given an encoding. \textbf{Right:} In TSDA, the discriminator classifies the domain index while the taxonomist reconstructs the domain taxonomy given encodings of data from different domains. }
\label{fig:da_vs_tsda}
\vskip -0.55cm
\end{figure}

\textbf{Adversarial Domain Adaptation.} 
There is a rich literature on domain adaptation~\cite{TLSurvey,MMD,CDAN,MCD,GTA,MDD,moment,blending,domain_bridge,DA_distillation, Zou_2018_ECCV, kumar2020understanding,prabhu2021sentry,maria2017autodial,mancini2019adagraph, tasar2020standardgan,jin2022domain}. 
% Domain adaptation has received much attention accompanied by notable prior work
To adapt a model across domains, existing methods typically align the encoding distributions of source and target domains, either by {direct matching} ~\cite{MMD,DDC,CORAL,moment,DA_distillation,CIDA} or {adversarial training}~\cite{DANN,CDANN,ADDA,MDD,UDA-SGD,blending,domain_bridge}. The adversarial domain adaptation framework becomes increasingly popular recently {due} to its {solid} theoretical foundation~\cite{GAN,ben2010theory,AMSDA,redko2020survey,MDD,InvariantDA}, {efficient} end-to-end implementation, and promising performance. Generally, these methods train an encoder to produce domain-invariant {representations by} trying to {fool} a discriminator that is trained to distinguish different domains; essentially they aim to perfectly align data from different domains in the encoding space, i.e., uniform alignment. Such uniform alignment can sometimes harms domain adaptation performance because it completely remove useful domain similarity information (e.g., information captured by a domain taxonomy) from the encoding. 
In contrast, our TSDA relaxes uniform alignment by introducing a taxonomist as the fourth player to recover the domain taxonomy from encodings (as shown in~\figref{fig:da_vs_tsda}), thereby significantly improving adaptation performance. 

\textbf{Domain Adaptation Related to Taxonomies.}  
Loosely related to our method are works related to both domain adaptation and taxonomies. 
For instance, \citet{TACS} \footnote{This paper has another version \cite{gong2021tada}. We cited the official published version.} focuses on the \emph{label} taxonomies and considers domain adaptation on semantic segmentation with one source domain and one target domain; it assumes the source domain and the target domain have different label taxonomies and aims to alleviate label shift~\cite{wu2019domain} between these two domains. In contrast, TSDA considers a \emph{completely different} setting. Specifically, \citet{TACS} adapts between \emph{two domains} with different label distributions, where taxonomies are used to describe \emph{labels relations}. Our TSDA adapts across \emph{multiple domains} (e.g., with each species as a domain) according to a domain taxonomy (e.g., an animal taxonomy), where taxonomies are used to describe \emph{domain relations}. \citet{TACS} is therefore \emph{not applicable} to our setting. Also related to our work is graph-relational domain adaptation (GRDA)~\cite{GRDA}, which adapts across domains connected by a graph with each domain as a node, continuously indexed domain adaptation (CIDA)~\cite{CIDA}, which adapts across continuously indexed domains, and variational domain indexing (VDI)~\cite{VDI}, which infers domain indices in CIDA using a hierarchical Bayesian deep learning model~\cite{CDL,BDL,BDLThesis,RPPU,BIN,BDLSurvey,ZESRec}. While a domain taxonomy can be reduced to a graph or continuous indices, it loses important hierarchical information and therefore often leads to suboptimal performance; this is verified by our empirical results in~\secref{sec:expr}.

\section{Method} \label{sec:method}
\subsection{Problem Setting and Notation}

% GH's version (edited by HW)
In domain adaptation, an input $\x \in \gX$ (e.g., a bird image) is used in conjunction with an additional domain specification {$u \in \gU = \{1,2,\dots,N\}$} (e.g., the species) to predict a label $y \in \gY$ (e.g., the wing color). Here we consider an unsupervised domain adaption setting {with $N$ domains}, where labeled data $\{(\x_l^s,u_l^s, y_l^s)\}_{l=1}^n \subseteq \gX \times \gU_s \times \gY$ are only available in the \emph{source} domains $\gU_s \subseteq \gU$, and the goal is to make predictions for the unlabeled data $\{(\x_l^t,u_l^t)\}_{l=1}^m \subseteq \gX \times \gU_t$ in the \emph{target} domains $\gU_t \subseteq \gU$. Note that we do not require $\gU_s \cap \gU_t = \emptyset$. {The goal is to predict the target-domain labels $\{(y_l^t)\}_{l=1}^m$ given source-domain labeled data and target-domain unlabeled data.}

In this work, we assume that an additional taxonomy $\gT$ is given; $\gT$ specifies a hierarchical similarity structure over the domains $\gU$. Formally, a taxonomy $\gT$ can be represented as a (directed) tree: 
\begin{itemize}
\vspace{-1.5mm}
\setlength\itemsep{-0.2em}
  \item The root node is $\gU$.
  \vspace{-0.5mm}
  \item In each layer, a parent $\gU_p \subseteq \gU$ is split into disjoint child nodes $\gU^p_{1},...,\gU^p_{n_p}$, where $\gU^p_{1} \cup ... \cup \gU^p_{n_p} = \gU_p$.
  \vspace{-0.5mm}
  \item Each leaf node only contains one domain $\{u\}$.
\vspace{-1.5mm}
\end{itemize}
We emphasize that a taxonomy of domains is different from a tree of domains, as a non-leaf node in a taxonomy does not (necessarily) correspond to a single domain. 

\subsection{Taxonomy-Structured Domain Adaptation}\label{sec:tsda}
\textbf{Overview.} 
{To balance representation learning between domain similarity and domain invariance,} we develop a game-theoretic formulation with four players: 1) an encoder $E$ that aims to produce the desired representation, 2) a discriminator $D$ that prevents the encoder from picking up domain dependencies, 3) a taxonomist $T$ that encourages the representation to preserve similarity information, and 4) a predictor $F$ that produces predictions according to the encoder and therefore guides the representation to strike a balance for the prediction task of interest. 

{Before further elaboration on the game, note that one key challenge in the framework is to model the taxonomy--the key object representing similarity among domains. Indeed, taxonomy is a combinatorial object that seems naturally incompatible with the continuously distributed representations modeled by deep networks. To bridge the gap, we transform the taxonomy $\gT$ to a \emph{distance matrix} $\mA$, where $\mA_{ij}$ records the shortest distance between two domains $i$ and $j$ on the taxonomy $\gT$.} Below, we formally define the game. 

\textbf{Encoder.} The encoder $E$ leverages all the available information to build a representation $\e_l$. It takes as input the data $\x_l$, domain index $u_l$, and the distance matrix $\A$. To facilitate learning, we first leverage an intermediate domain embedding $\z_{u_l} = g(u_l, \A)$ to capture the immediate dependency between $u_l$ and $\mA$: 
\begin{equation*}
\e_l = E(\x_l,u_l,\A) = f(\x_l,g(u_l,\A)) = f(\x_l,\z_{u_l}),
\end{equation*}
where the encoder $E(\cdot)$ is defined by composition of $g(\cdot)$ and $f(\cdot)$. 
% Theoretically, any domain embedding $\z_{u_l}$ works equally well as long as it forms a bijection to the domain identity set $\gU$ (e.g., one-hot vectors). 
{In principle, given sufficiently powerful $f$, every embeddings $\z_{u_l}$ works equally well as long as a bijection to the set of domains $\gU$ can be formed. Here we use a simple pretraining procedure to obtain the domain embedding based on a reconstruction loss with respect to the domain distance matrix $\mA$ ({see more details in the Appendix}).} 

\textbf{Discriminator}. The discriminator $D$ aims to identify the domain from the encoding $\e_l$. This is realized as a function $D: \gZ \to \gU$ that minimizes a domain identification (classification) loss $l_d$: 
\begin{equation*}
L_d(D,E)  \triangleq \mathbb{E}[l_d(D(E(\x_l,u_l,\A),u_l)], 
\end{equation*}
where the expectation $\mathbb{E}$ is taken over the data distribution $p(\x,u)$; $l_d$ is typically realized by cross-entropy. 

\textbf{Taxonomist.} Similar to the discriminator $D$, the taxonomist $T$ aims to recover taxonomy information within the encoding $\e_l$. Unlike the vanilla categorical domain index that can be easily compared element-wisely, the distance matrix $\mA$ implied by the taxonomy is pairwise in nature. We therefore specify the taxonomy with paired inputs $T: \gZ \times \gZ \to \mathbb{R}$ and optimizes it with respect to the distance matrix $\mA$:
\begin{align*}
L_t(T,E)  \triangleq \mathbb{E} [l_t(&T(E(\x_1,u_1,\A), E(\x_2,u_2,\A)), \A_{u_1,u_2})], 
\end{align*}
where the expectation $\mathbb{E}$ is taken over a pair of i.i.d. samples $(\x_1, u_1),(\x_2, u_2)$ from the joint data distribution $p(\x, u)$. Here $l_t$ denotes a regression loss (e.g., $\ell_2$ distance). 
% \blue{
% }

\textbf{Predictor}. %Rather than examining information, 
The predictor $F: \gZ \to \gY$ simply takes encoding $\e_l$ as input and outputs a label. We therefore optimize it with respect to the labels:
\begin{equation*}
L_f(F,E)  \triangleq \mathbb{E}^s [l_f(F(E(\x_l,u_l,\A),y_l)], 
\end{equation*}
where the expectation $\mathbb{E}^s$ is taken over only the data in the source-domain data distribution $p^s(\x,y,u)$, since labels are not available in the target domains. Here $l_f$ could be specified according to the prediction task at hand (e.g., cross-entropy loss for classification or $\ell_2$ distance for regression). 

\textbf{The Full Game}. With the aforementioned players, we are now ready to specify the full game. 
\begin{equation}
\min_{E,F,T} \max_D L_f(E,F)-\lambda_d L_d(D,E)+\lambda_t L_t(T,E),\label{eq:min_max_game}
\end{equation}
where $E$, $F$, and $T$ play cooperatively against the discriminator $D$. The opposite optimization direction between the discriminator $D$ and the taxonomist $T$ is due to the inherent difference in their inducing properties: the discriminator $D$ aims to find remaining domain information in the encoding, thus in an adversarial position with respect to the encoder; in contrast, the encoder and the taxonomist have the same goal of keeping taxonomy information within the encoding. The hyper-parameters $\lambda_d, \lambda_t \geq 0$ are introduced to enable a flexible trade-off, such that the encoder $E$ and the predictor $F$ can be learned with proper regularization.

{\textbf{Discussion.}} Traditional adversarial domain adaptation enforces the encoder $E$ to fool the discriminator $D$, so that it aligns all domains uniformly. In our model, due to the addition of taxonomist $T$, the encoder $E$ has to retain a certain amount of domain information in order to recover the taxonomy $\A$, such that the alignment is no longer uniform. As a result, the discriminator must compete with the taxonomist during the optimization process to reach an optimal balance, thus adapting successfully across domains in the taxonomy. Detailed analysis of the competition would be done in \secref{sec:theory}.

\section{Theory} \label{sec:theory}
% \secref{sec:method} hightlights the intuition for introducing taxonomist $T$ into our model. This intuition motivates us to further analyze the min-max game for . 

In this section, we will prove the intuition mentioned in \secref{sec:method}. With the addition of the taxonomist $T$, the discriminator cannot enforce perfect alignment on encoding space. An interesting corollary is that TSDA can recover DANN with a non-informative taxonomy, highlighting the flexibility of our model. Furthermore, we will also discuss a straightforward extension of DANN with weighted pairwise discriminators. We prove that such DANN only produces uniform alignment, and therefore cannot incorporate the taxonomy information during adaptation.

\subsection{Analysis of the Taxonomist}\label{sec:tax}

% \ghl{new below}

In this section, we formally show that the proposed formulation indeed allows the model to achieve a balance between the two contrasting goals: removing domain information and preserving domain structure. Here we focus on the analysis of the encoder which is the direct subject of the regularization. 
% Following prior literature, we say that domain information is \emph{fully removed}\footnote{This is often called ``uniformly aligned'' in the literature. We use ``fully removed'' to highlight its difference from preserving domain taxonomy.} if $\e \indep u$, {which means the encoding distributions of every domain are aligned $p(\e|u=1)=p(\e|u=2)=\dots=p(\e|u=N)$.} This is also called \emph{uniform alignment}~\cite{GRDA}, as defined below. 
{Following prior literature, we say that domain information is \emph{fully removed} or the domains are \emph{uniformly aligned} if $\e \indep u$, {which means the encoding distributions of every domain are aligned as defined below.} }

\begin{definition}[\textbf{Uniform Alignment}]\label{def:ua}
% A domain adaptation model achieves uniform alignment if its encoder $\e=E(\x)$ ensures that $p(\e|u)=p(\e), \forall u$. 
A domain adaptation model achieves uniform alignment if its encoder $\e=E(\x)$ {satisfies} $p(\e|u)=p(\e), \forall u$. 
\end{definition}

On the other hand, the taxonomy information is \emph{fully retained} if $L_t(T, E) = 0$, {i.e., the domain taxonomy can be perfectly reconstructed from the encodings}. 

We begin our analysis by showing that the two goals are contradictory except for the scenario where the taxonomy does not contain any extra information beyond the domain index $u$; We say that such a domain taxonomy is \emph{non-informative}, with the formal definition below. 

\begin{definition}[\textbf{Non-Informative Domain Taxonomy}]
A domain taxonomy is \emph{non-informative} if and only if $\A_{ij} = a, \forall i \neq j$, for some constant $a \in \sZ_{>0}$, where $\A$ is the domain taxonomy's associated domain distance matrix. 
\end{definition}
% ; this happens when $\A_{ij} = a, \forall i \neq j$, for some constant $a \in \sZ_{>0}$. 
In words, if the distance of every pair of domains is the same, the taxonomy structure cannot provide meaningful comparisons of the similarity between different domains. 
% ; we call such a domain taxonomy \emph{non-informative}. 
A domain taxonomy is non-informative if all the domains have the same parent. Now we present the main result. %{(proof in the Appendix).}
% Formally, we have
\begin{theorem}[\textbf{Incompatibility}]\label{thm:txn_value}
If $\e \indep u$ and $\ell_2$ distance is used in $L_t(\cdot)$, 
\begin{align*}
    L_t(T,E) = 0 \implies \A_{ij} = a, \forall i \neq j \in \gU
\end{align*}
for some $a \in \sZ_{>0}$. 
\end{theorem}
% Here we make a few remarks before presenting the proof. 
{We make a few remarks here.}
First, {the theorem essentially states that if the domain taxonomy is not non-informative, preserving taxonomy information and achieving uniform alignment is \emph{incompatible}; that is, the taxonomy information cannot be fully preserved if uniform alignment is achieved (i.e., the domain information is fully removed). Second, the other direction}
\begin{align}
    \A_{ij} = a, \forall i \neq j \in \gU \implies \min\nolimits_{T} L_t(T,E) = 0 \label{eq:trivial}
\end{align}
{holds trivially without requiring $\e \indep u$ so long as $T$ is not less powerful than a constant function. Note that the theorem is stronger than \eqnref{eq:trivial} since no optimization is involved. Albeit the intuitive nature of the statement, the proof is not trivial (see the Appendix for details). 
}
% holds trivially without requiring $\e \indep u$ so long as $T$ is not less powerful than a constant function. Note that the theorem is stronger than Eq.~(\ref{eq:trivial}) since no optimization is involved. Second, {the theorem essentially states} that if the domain taxonomy is not non-informative, the taxonomy information cannot be fully preserved if the domain information is fully removed.  Albeit the intuitive nature of the statement, the proof in not trivial. 

% \begin{proof}
% \begingroup\makeatletter\def\f@size{9}\check@mathfonts
% We first expand the loss as
% \begin{align*}
%      L_t(T,E) = & \frac{1}{2}\int\!\!\!\!\int \sum_{i \neq j} (\A_{ij} \!-\! T(\e, \e'))^2 p(\e, i) p(\e', j)\ d\e \, d\e'\\
%               = & \frac{1}{2N^2} \int\!\!\!\!\int \sum_{i \neq j} (\A_{ij} - T(\e, \e'))^2 p(\e) p(\e') \ d\e \, d\e' \\
%               = & \frac{1}{2N^2} \sum_{i \neq j} \A_{ij}^2 -2\A_{ij}\mathbb{E}[T(\e, \e')] + \mathbb{E}[T(\e, \e')^2], 
% \end{align*}
% where the second equality is due to $p(\e, i) = p(\e) p(i) = p(\e) / N$. We also have the equality
% \begin{align*}
%      \mathbb{E}[T(\e, \e')^2] = \mathbb{E}[T(\e, \e')]^2 + \text{Var}[T(\e_i, \e_j)].
% \end{align*}
% Combining the two {equalities}, we get

% \begin{align*}
% L_t(T,E) = & \alpha \text{Var}[T(\e, \e')] + \beta \sum_{i \neq j} [\A_{ij} - \mathbb{E}[T(\e, \e')]]^2 
% \geq 0,
% \end{align*}
% where $\alpha = (N)(N-1)/(2N^2)$ and $\beta = 1/(2N^2)$. Finally,
% \begin{align*}
% L_t(T,E) = 0 \implies \A_{ij} =  \mathbb{E}[T(\e, \e')], \forall i \neq j,
% \end{align*}
% completing the proof. 
% \endgroup
% \end{proof}

{We also show that TSDA can recover DANN as a special case in \crlref{cor:tda2dann} (proof in the Appendix); therefore TSDA is methodologically more general than DANN.}

\begin{cor}[\textbf{TSDA Generalizes DANN}]\label{cor:tda2dann}
Omitting the predictor, if the taxonomy is non-informative, then the optimum of TSDA is achieved if and only if the embedding distributions of all the domains are the same, i.e. $p(\e|u=1)=\cdots=p(\e|u=N)=p(\e), \forall \e$.
\end{cor}

\begin{figure*}[!t]
\vskip -0.3cm
\begin{center}
% \centerline{\includegraphics[width=\columnwidth *2]{pic/expr1.pdf}}
\centerline{\includegraphics[width=0.99\textwidth]{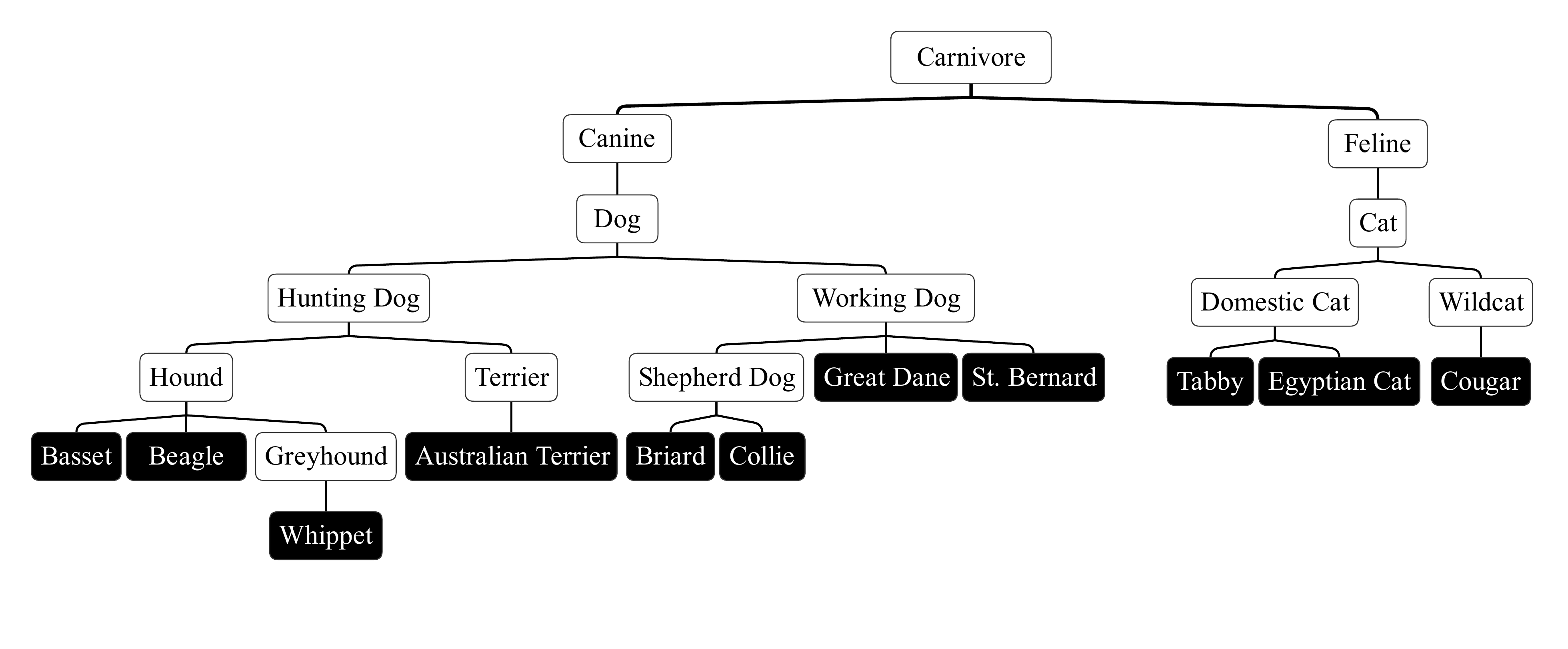}}
\vskip -0.45in
\caption{Domain taxonomy of \emph{ImageNet-Attribute-DT} with $11$ domains shown as leaf nodes. Note that leaf nodes are marked with black base and white text. Non-leaf nodes and their connection to leaf nodes are obtained from WordNet \cite{wordnet}.  }%Non-leaf nodes are obtained according to WordNet\cite{wordnet}. 
\label{fig:im}
\end{center}
\vskip -0.4in
\end{figure*}

\subsection{Effective Hyperparameters}

{As discussed in~\secref{sec:tsda}, $\lambda_d$ and $\lambda_t$ of~\eqnref{eq:min_max_game} balance 1) the encoder-discriminator sub-game, which tries to achieve uniform alignment by removing domain-specific information in the representation $\e$, and 2) the encoder-taxonomist sub-game, which tries to preserve the part of domain-specific information captured by the domain taxonomy. We show in \thmref{thm:lambda_ua} that their ratio ${\lambda_t}/{\lambda_d}$ needs to be large enough to make the balancing effect to happen.

\begin{theorem}[\textbf{Uniform Alignment, $\lambda_t$, and $\lambda_d$}]\label{thm:lambda_ua}
If $\lambda_t > \lambda_d$ and the
domain taxonomy is not non-informative, 

$\min_{E, T} \max_D -\lambda_d L_d(D,E)+\lambda_t L_t(T,E)$ will not yield uniform alignment (\defref{def:ua}). \end{theorem}

\thmref{thm:lambda_ua} implies that we need $\lambda_t > \lambda_d$ to prevent TSDA from converging to a trivial solution where the taxonomist is ignored and uniform alignment is achieved. In such a trivial solution, no information on the domain taxonomy is preserved, and TSDA degenerates to DANN-like methods. Guided by \thmref{thm:lambda_ua}, $\lambda_t$ and $\lambda_d$ in our experiments are chosen such that $\lambda_t >\lambda_d$ (more details in~\secref{sec:expr}). 

\subsection{An Alternative Method and Its Analysis}

Since TSDA involves an extra taxonomist, one might wonder whether the same effect can be achieved by simply adjusting existing methods without algorithmic innovation. Here we show that this can be done easily by showing an impossibility result for a natural extension of DANN for domain taxonomy. Note that DANN is equivalent to TSDA without the taxonomist (i.e., $\lambda_t = 0$). 

To model the pairwise distance induced by a domain taxonomy, a straightforward way is to utilize a weighting function $w_{ij}$, which captures the distance for each pair of domains. For example, the weight could be inversely proportional to the domain distance $w_{ij} \propto 1/\A_{ij}$. We can then adapt DANN by utilizing a distinct discriminator $D_{ij}$ for each pair of domains $(i, j)$: 
\begin{align}
    \min_E \max_{D_{ij}} \mathbb{E}[\sum_{j \neq i} w_{ij} \log D_{ij} (E(\x)) ], \label{eq:pair}
\end{align}
where the expectation is over the data distribution $p(\x, u)$. For clarify, here we omit the impact of the predictor to simplify the analysis. Note that for every encoder $E$, each inner-maximization problem for $D_{ij}$ reduces to a standard adversarial domain adaptation problem with two domains. We therefore can easily invoke an existing {result~\cite{DANN,GRDA}} to find the optimal solution. 

\begin{lemma}[\textbf{Optimal Discriminator}]\label{lem:opt_dis_dannafdsadfsa}
For every $E$, the optimal $D_{ij}$ of Eq.~(\ref{eq:pair}) satisfies
\begin{align*}
    D_{ij}(\e) =  \frac{p(\e|u=i)}{p(\e|u=i) +p(\e|u=j)}, \forall \e \in \gZ.
\end{align*}
\end{lemma}
That says, for each pair $(i, j)$ of domains, the optimal discriminator simply uses the appearing ratio of $\x$ under the two domains as the output. We can then use this result to approach the equilibrium of the minimax game (\eqnref{eq:pair}). 

\begin{theorem}[\textbf{{Optimal Encoder}}]\label{thm:opt_value_tsda}
%Given an adjacency matrix $\A$, 
The min-max game in \eqnref{eq:pair} has a tight lower bound:
\begin{align*}
% L_{d'}(D, E) \geq \frac{\log 2}{N} \sum_{i\neq j} w_{ij},
\max_{D_{ij}} \mathbb{E}[\sum_{j \neq i} w_{ij} \log D_{ij} (E(\x)) ]  \geq -\frac{\log 2}{N} \sum_{i\neq j} w_{ij},
\end{align*}
where $N$ denotes the number of domains. Furthermore, the equality, i.e., the optimum, is achieved when
\begin{align*}
 p(\e|u=i) = p(\e|u=j),\mbox{ for any } i,j,
\end{align*}
or equivalently, $p(\e|u=i) = p(\e)$.
\end{theorem}

\thmref{thm:opt_value_tsda} states that regardless of the value for $w_{ij}$, encoder will always produce uniform alignment. This indicates such a modified DANN failed to incorporate taxonomy information into the domain adaptation process. {The proof is available in the Appendix.}

\section{Experiments}
\label{sec:expr}
In this section, we evaluate TSDA and existing methods on both synthetic and real-world datasets.

\begin{figure*}[!t]
\vskip -0.12in
\begin{center}
% \centerline{\includegraphics[width=\columnwidth *2]{pic/expr1.pdf}}
\centerline{\includegraphics[width=0.99\textwidth]{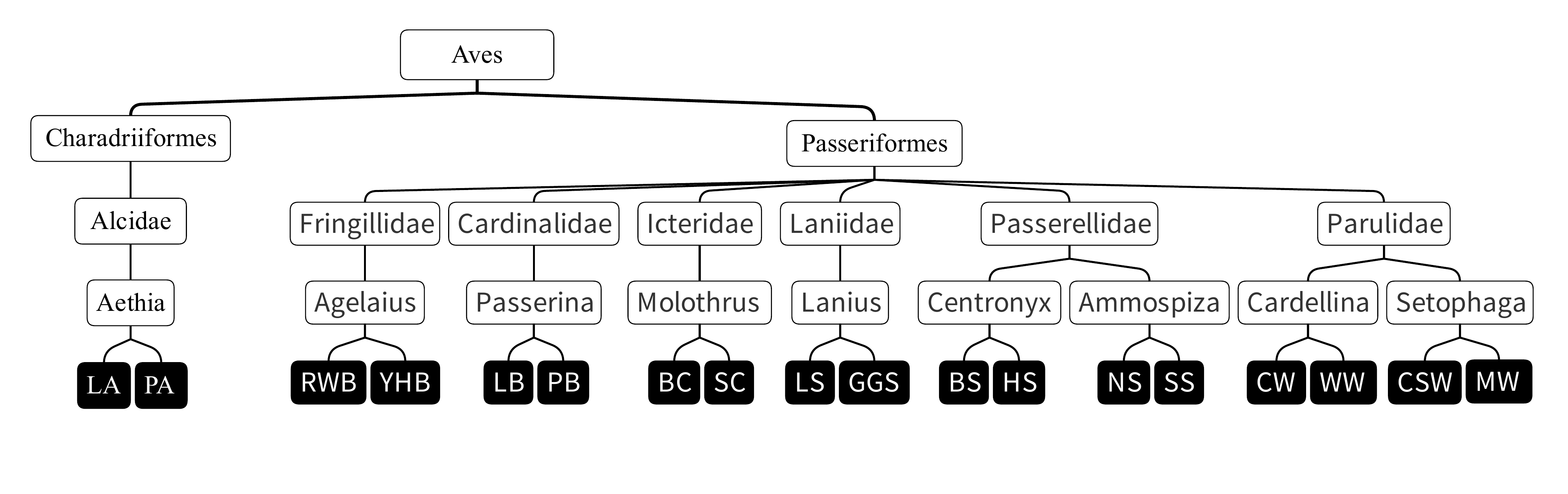}}
\vskip -0.45in
\caption{Domain taxonomy of \emph{CUB-DT} with $18$ domains shown as leaf nodes. Note that leaf nodes are marked with black base and white text. 
% Non-leaf nodes and their connection to leaf nodes are obtained from WordNet\cite{wordnet}. 
% Domain taxonomy of \emph{CUB-DT} with $18$ domains for adaptation all shown as leaf nodes. Note that leaf nodes are marked with black base and white text Non leaf nodes are obtained according to real world taxonomy rules. All nodes are in scientific name in which leaf nodes has abbreviation of common names. 
For clarity we abbreviate domain names in the figure: Least Auklet (\textbf{LA}), Parakeet Auklet (\textbf{PA}), Red Winged Blackbird (\textbf{RWB}), Yellow Headed Blackbird (\textbf{YHB}), Lazuli Bunting (\textbf{LB}), Painted Bunting (\textbf{PB}), Bronzed Cowbird (\textbf{BC}), Shiny Cowbird (\textbf{SC}), Loggerhead Shrike (\textbf{LS}), Great Grey Shirke (\textbf{GGS}), Baird Sparrow (\textbf{BS}), Henslow's Sparrow (\textbf{HS}), Nelson's Sparrow (\textbf{NS}), Seaside Sparrow (\textbf{SS}), Canada Warbler (\textbf{CW}), Wilson's Warbler (\textbf{WW}), Chestnut-sided Warbler (\textbf{CSW}), Myrtle Warbler (\textbf{MW}). }
\label{fig:cub}
\end{center}
\vskip -0.25in
\end{figure*}

\begin{figure*}[!t]
\vskip 0cm
\begin{center}
% \centerline{\includegraphics[width=\columnwidth *2]{pic/expr1.pdf}}
\centerline{\includegraphics[width=0.99\textwidth]{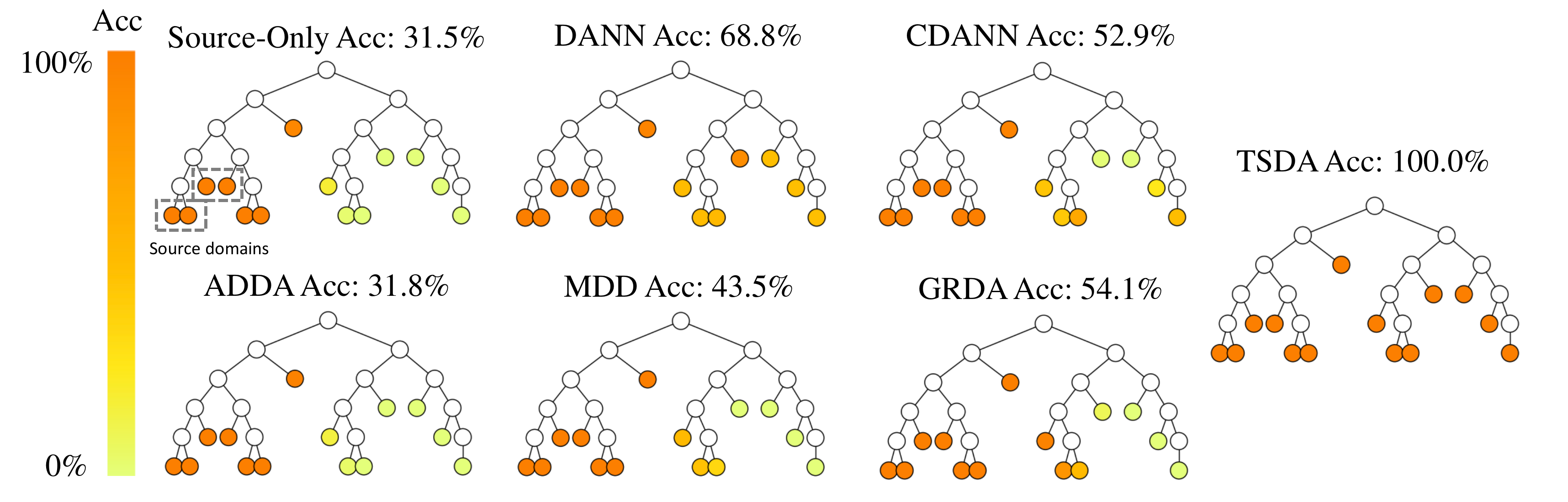}}
\vskip -0.22in
\caption{Detailed results on \emph{DT-14} with $14$ domains. We use the $4$ domains in the dashed box as source domains. The spectrum from `orange' to `yellow' indicates accuracy from $100\%$ to $0\%$ (best viewed in color). }%Source-Only's accuracy on \emph{DT-14} is $31.5\%$; we omit Source-Only in the figure to prevent clutter. 
\label{fig:dt-14_acc}
\end{center}
\vskip -0.35in
\end{figure*}

\subsection{Datasets}
\emph{\textbf{DT-14}} is a synthetic binary classification dataset with 14 domains, each consisting of $100$ positive and negative labeled data points.
To simulate real-world scenarios, we argue that a synthetic dataset should have an informative domain taxonomy (as defined in \secref{sec:tax}) which reflects the similarity among domains. 
% To meet these requirements, we divide our generation into 2 steps: first, we generate a perfect tree that nodes with nearest common ancestors have closer feature vectors. Then we prune this tree to reduce the number of leaf nodes, which gives us an informative tree. Notice that in the first step, every node (including the internal nodes) will be assigned a feature vector, because after pruning, every node is possible to become the new leaf node. Our synthetic domains lies in the location of the new leaf nodes, and the feature vectors on nodes are then employed to generate the decision boundary for each domain. Specifically, we use a bottom-up method to generate a perfect binary tree with required feature vectors. 
We first randomly generate $2^5=32$ unit vectors $[a_i, b_i]$, ($a_i \in \mathbb{R}, b_i \in \mathbb{R^+}$)
% with $b_i \geq 0$
and denote their angles as $\theta_i = \arcsin(\frac{b_i}{a_i})$. We sort all the unit vectors by their angles, and pair consecutive unit vectors according to this order (e.g., $[a_0, b_0]$ with $[a_1, b_1]$, $[a_2, b_2]$ with $[a_3, b_3]$). We then assign each pair a ``parent" unit vector $[a'_i, b'_i] = \frac{1}{2} [a_{2i} + a_{2i+1}, b_{2i} + b_{2i+1}]$. This leads to a new group of unit vectors, and we can repeat the previous steps until we reach the ``root'' node. 
This produces a 6-level unit vector tree (a perfect binary tree) with $32$ leaf nodes. 
We then randomly prune the tree to generate the final domain taxonomy with $14$ leaf nodes (\figref{fig:dt-14_acc}), each associated with a unit vector. 
% After reducing the number of leaf nodes to 14 by random pruning, we then generate data for each domain. 
We randomly generate positive data $(\x, i, 1)$ and negative data $(\x, i, 0)$ from two different $2$-dimensional Gaussian distributions, $\mathcal{N}(\muu_{i,1}, \I)$ and $\mathcal{N}(\muu_{i,0}, \I)$ where $\muu_{i,1} = [\frac{\omega_i}{\pi}a_i, \frac{\omega_i}{\pi}b_i]$ and $\muu_{i,0} = [-\frac{\omega_i}{\pi}a_i, -\frac{\omega_i}{\pi}b_i]$. See \figref{fig:dt-14_acc} for the generated domain taxonomy. 

% We argue that to simulate the real-world scenario, a synthetic dataset should 

% To ensure that domains at same taxonomy level share similar 

% We simulate the real-world domain taxonomy by randomly pruning a perfect tree. Specifically, we first randomly generated 32 unit vector $[a_i, b_i]$, with vector angle $\theta_i = \arcsin(\frac{b_i}{a_i})$.

\begin{table*}[!tbhp]
\vskip -0.1cm
\centering
% \begin{scriptsize}
  \caption{Accuracy for each of the $6$ target domains on the \emph{ImageNet-Attribute-DT} dataset (domain taxonomy in~\figref{fig:im}) as well as the average accuracy for different methods. Note that there is only one single DA model per column. We mark the best result with \textbf{bold face}.}
  \label{tab:im}
  \vskip 0.1cm
  \centering
  \begin{tabular}{lcccccccc} \toprule
      Target Domain           & Source-Only & DANN & CDANN & ADDA & MDD & GRDA & TSDA \\ \midrule
         Basset               & 84.0 & 84.0  &  72.0   &  88.0  &  88.0 &  84.0  &  \textbf{92.0}\\
         Beagle               & 68.0 & 64.0  &  68.0   &  44.0  &  68.0 &  \textbf{76.0}  &  \textbf{76.0}\\
         Whippet              & 68.0 & 64.0  &  68.0   &  68.0  &  \textbf{76.0} &  72.0  &  \textbf{76.0}\\
         Australian Terrier & 80.0 & 80.0  &  72.0   &  84.0  &  \textbf{84.0} &  \textbf{84.0}  &  \textbf{84.0}\\
         Briad                & \textbf{80.0} & \textbf{80.0}  &  \textbf{80.0}   &  \textbf{80.0}  &  72.0 &  68.0  &  72.0\\
         Collie               & 84.0 & 80.0  &  \textbf{88.0}   &  84.0  &  84.0 &  84.0  &  84.0\\ \midrule
         Average           & 77.3 & 75.3 & 74.7  & 74.7 & 78.7 & 78.0  &  \textbf{80.7}\\
     \bottomrule
  \end{tabular}
% \end{scriptsize}
\vskip -0.3cm
\end{table*}

\begin{table*}[!thp]
\vskip -0.1cm
\centering
% \begin{scriptsize}
  \caption{Accuracy for each of the $9$ target domains on the \emph{CUB-DT} dataset (domain taxonomy in~\figref{fig:cub}) as well as the average accuracy for different methods. Note that there is only one single DA model per column. We mark the best result with \textbf{bold face}. }
  \label{tab:cub}
  \vskip 0.1cm
  \centering
  \begin{tabular}{lcccccccc} \toprule
      Target Domain           &Source-Only & DANN & CDANN & ADDA   &    MDD   & GRDA & TSDA \\ \midrule
         Great Grey Shrike    & \textbf{95.0} &  78.3  &  73.3   &  58.3  &  80.0    &  23.3  &  \textbf{95.0}\\
         Baird Sparrow        & \textbf{71.7} &  28.3  &  35.0   &  50.0    &  40.0    &  63.3  &  53.3\\
         Henslow's Sparrow    & 65.0 &  48.3  &  50.0   &  63.3  &  58.3    &  \textbf{75.0}  &  61.7\\
         Nelson's Sparrow     & 80.0 &  73.3  &  70.0   &  86.7  &  86.7    &  80.0  &  \textbf{100.0}\\
         Seaside Sparrow      & 70.0 &  93.3  &  75.0   &  \textbf{96.7}  &  95.0    &  93.3  &  95.0\\
         Canada Warbler       & 76.7 &  70.0    &  70.0   &  75.0    &  80.0    &  60.0  &  \textbf{88.3}\\
         Wilson's Warbler     & 76.7&  73.3  &  63.3 &  66.7  &  78.3    &  41.7  &  \textbf{85.0}\\
         Chestnut-sided Warbler& 81.7 &  71.7 &  70.0   &  86.7  &  86.7    &  76.7  &  \textbf{93.3}\\
         Myrtle Warbler       & 66.7&  65.0    &  56.7 &  66.7  &  66.7    &  68.3  &  \textbf{70.0}\\ \midrule
         Average              & 75.9 & 66.9   & 62.6  & 72.2   & 74.6   & 64.6  &  \textbf{82.4}\\
         
     \bottomrule
  \end{tabular}
% \end{scriptsize}
\vskip -0.5cm
\end{table*}

\begin{table}[!t]
\small
\setlength{\tabcolsep}{2.2pt}
\vskip -0.0cm
% \begin{scriptsize}
  \caption{Accuracy (\%) on \emph{DT-14} and \emph{DT-40}. }
  \label{tab:dt}
  \vskip 0.1cm
  \centering
  \begin{tabular}{cccccccc} \toprule
   Method & SO & DANN & CDANN & ADDA & MDD & GRDA & {TSDA}\\\midrule
   \emph{DT-14} & 31.5 & 68.8 & 52.9 & 31.8 & 43.5 & 54.1 &\textbf{100.0} \\
   \emph{DT-40} & 43.1 & 55.4 & 43.4 & 43.1 & 42.9 & 44.0 & \textbf{82.6} \\

     \bottomrule
  \end{tabular}
%   \end{scriptsize}
  \vskip -0.cm
\end{table}

\emph{\textbf{DT-40}} is constructed with the same procedure as \emph{DT-14} except that it is pruned from a 7-level perfect binary tree, with 40 domains as leaf nodes after pruning and $b_i \in \mathbb{R}$. %Besides, we enlarge the value space for unit vector $[a_i, b_i]$, where $b_i$ now could be arbitrary. 
We select 6 domains as source domains, with others as target domains. % \hao{add pointer to fig 3?}

\emph{\textbf{ImageNet-Attribute-DT}}~\cite{imagenet_attribute} builds on the animal images from ImageNet, with additional attribute labels (e.g., whether the skin color is black or not). Here we focus on a binary classification task for the attribute ``brown", because this attribute is available for the largest number of image categories, i.e., $11$ categories. We use these $11$ image categories as $11$ domains, with ``Great Dane", ``St. Bernard", ``Tabby", ``Egyptian Cat" and ``Cougar" as source domains, and the others as target domains. Each domain contains $25$ images, and the domain taxonomy, shown in \figref{fig:im}, is constructed by the hierarchies of image categories from WordNet~\cite{wordnet}.

% contains over 400 object categories with 25 object attributes. 

% \emph{\textbf{ImageNet-Attribute-DT}} contains over 400 object categories with 25 object attributes verified through Amazon Mechanical Turk. Among these attributes, we focus on the binary classification task of the attribute ``brown", since ``brown'' is the attribute that is available in the largest number of domains. We select 11 domains as leaf nodes for adaptation and fill the rest of the domain taxonomy based on WordNet words relationship. 
% Each domain contains {25} images. 
% In such way, we construct a realistic graph without any external interference. We select 5 domains as source domains and use others as target domains. 

%\figref{fig:im}
% source domain: imagenet: 'Great Dane', 'St. Bernard', 'Tabby', 'Egpytian Cat', 'Cougar'。

% \hao{add pointer to fig 4? and specify which 5 domains are source}

\emph{\textbf{CUB-DT}}~\cite{cub200} contains $11{,}788$ images of $200$ bird categories. Every image is annotated with $312$ binary attributes (e.g., birds' body parts, shapes, colors). To ensure label balance, we choose to focus on the classification task of predicting ``whether the upper part of a bird is black''. We construct the domain taxonomy with $18$ domains based on the database of the National Center for Biotechnology Information (NCBI)~\cite{bird_database}, with ``LA", ``PA", ``RWB", ``YHB", ``LB", ``PB", ``BC", ``SC", ``LS" as sources domains and the others as target domains. These $18$ domains contain {$1{,}035$ images} in total. \figref{fig:cub} shows the constructed domain taxonomy. 

\subsection{Baselines and Implementation}
\label{subsec:implementation}
We compare TSDA with various state-of-the-art adversarial domain adaptation models, including Domain Adversarial Neural Networks (\textbf{DANN})\cite{DANN}, Adversarial Discriminative Domain Adaptation (\textbf{ADDA})\cite{ADDA}, Conditional Domain Adaptation Neural Networks (\textbf{CDANN})\cite{CDANN}, Margin Disparity Discrepancy (\textbf{MDD})\cite{MDD} and Graph-Relational Domain Adaptation (\textbf{GRDA})\cite{GRDA}. We also include results when one trains the model in the source domains is directly test it in the target domains (\textbf{Source-Only} or \textbf{SO} in short). 
Since each domain in \emph{ImageNet-Attribute-DT} and \emph{CUB-DT} represents a real-world category or species, we report both individual accuracy in each target domain and average accuracy over all target domains.
All models above are implemented in PyTorch. The balancing hyperparameters $\lambda_d$ and $\lambda_t$ range from $0.1$ to $1$ (see the Appendix for more details on training). For a fair comparison, the encoder for all the baselines takes as input the data $\x$, the domain index $u$, and the node embedding $\z$ (see the Appendix for more implementation details). % For \emph{ImageNet-Attribute-DT} and \emph{CUB-DT}, data $\x$ are embeddings extracted from pretrained Resnet-18.%\hao{use 'the Appendix' not Appendix}

\subsection{Results}
\textbf{\emph{DT-14} and \emph{DT-40}}. \tabref{tab:dt} shows the accuracy of all the baselines and TSDA on the synthetic datasets. In \emph{DT-14}, Source-Only ``overfits'' source domains and does not generalize to target domains, achieving an accuracy of $31.5\%$, significantly lower than random guess (50\%). 
Both DANN and CDANN only slightly outperform random guesses, with DANN as the best baseline. Interestingly, other baselines, ADDA, MDD, and GRDA even underperform random guesses. This is possible because these baselines either ignore the domain taxonomy and blindly align different domains (DANN, CDANN, ADDA, and MDD) or fail to faithfully capture the information in the domain taxonomy during adaptation (GRDA). In contrast, TSDA successfully performs domain adaptation by aligning data from different domains according to the domain taxonomy, thereby achieving significantly higher accuracy. Similarly in \emph{DT-40}, DANN is the best baseline and our TSDA significantly outperforms all baselines. Note that the accuracy in \emph{DT-40} is generally lower than that in \emph{DT-14} since \emph{DT-40} has more domains and a more complex domain taxonomy.

\figref{fig:dt-14_acc} shows the detailed accuracy of TSDA and all baselines in each domain of \emph{DT-14}. The spectrum from ``orange" to ``yellow" on the left indicates accuracy from $100\%$ to $0\%$. For all baselines, it is clear that target domains that are closer to source domains tend to have higher accuracy. 
This is expected because adjacent domains have similar decision boundaries, and therefore traditional adversarial DA methods are able to achieve reasonable accuracy in nearby domains by blindly enforcing uniform alignment. However, their performance is substantially worse for target domains farther away from source domains. In contrast, TSDA achieves promising results in all domains. 

\begin{figure}[!t]
% \vskip -0.8cm
\begin{center}
% \centerline{\includegraphics[width=\columnwidth *2]{pic/expr1.pdf}}
\centerline{\includegraphics[width=0.48\textwidth]{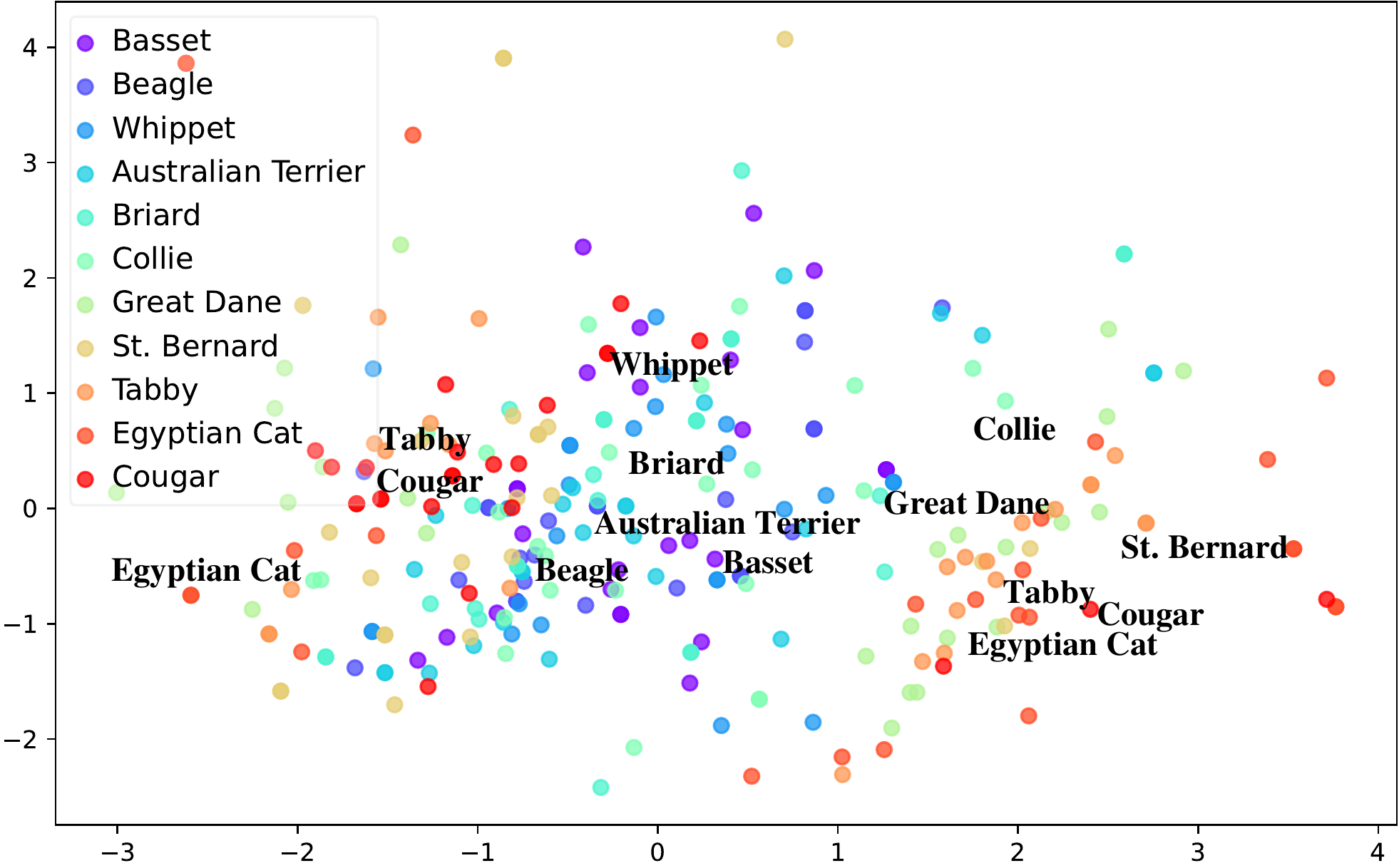}}
\caption{TSDA's learned encoding $\e$ for $11$ domains on \emph{ImageNet-Attribute-DT}. Domains related to ``dogs'', e.g., ``Basset'' and ``Beagle'', contain encodings in the middle, while domains related to ``cats'', e.g., ``Tabby'', contain encodings on both sides; this is consistent with the domain taxonomy in~\figref{fig:im}. Each domain contains $2$ clusters because of the binary classification task. }
\label{fig:im_vis}
\end{center}
\vskip -0.35in
\end{figure}

\textbf{\emph{ImageNet-Attribute-DT}}. \tabref{tab:im} shows the accuracy for each of the $6$ target domains on \emph{ImageNet-Attribute-DT} as well as the average accuracy for different methods. Compared to Source-Only, DANN, CDANN, and ADDA achieve a negative performance boost in terms of average accuracy, demonstrating the difficulty of performing DA across taxonomy-structured domains. Both MDD and GRDA slightly outperform Source-Only in terms of average accuracy; however, MDD's and GRDA's accuracy falls under $70\%$ in domain ``Beagle'' and domain ``Briad'', respectively. In contrast, our TSDA manages to outperform all baselines in terms of average accuracy and achieve accuracy higher than $70\%$ in every individual target domain. 

\figref{fig:im_vis} plots TSDA's learned encodings $\e$ on all $11$ domains of \emph{ImageNet-Attribute-DT}. %We color the encodings according to their associated domains. 
Interestingly, the learned encodings' positions are consistent with the domain taxonomy in~\figref{fig:im}. For example, domains related to ``dogs'', e.g., ``Basset'' and ``Beagle'', contain encodings in the middle of \figref{fig:im_vis}, while domains related to ``cats'', e.g., ``Egyptian Cat'', contain encodings on both sides of \figref{fig:im_vis}. Moreover, domain ``Basset'' and domain ``Beagle'' share the same parent in the domain taxonomy, and encodings from these two domains are close in \figref{fig:im_vis}. Note that each domain contains $2$ clusters because of the binary classification task.

\begin{figure}[!t]
% \vskip 0.1cm
\begin{center}
% \centerline{\includegraphics[width=\columnwidth *2]{pic/expr1.pdf}}
\centerline{\includegraphics[width=0.49\textwidth]{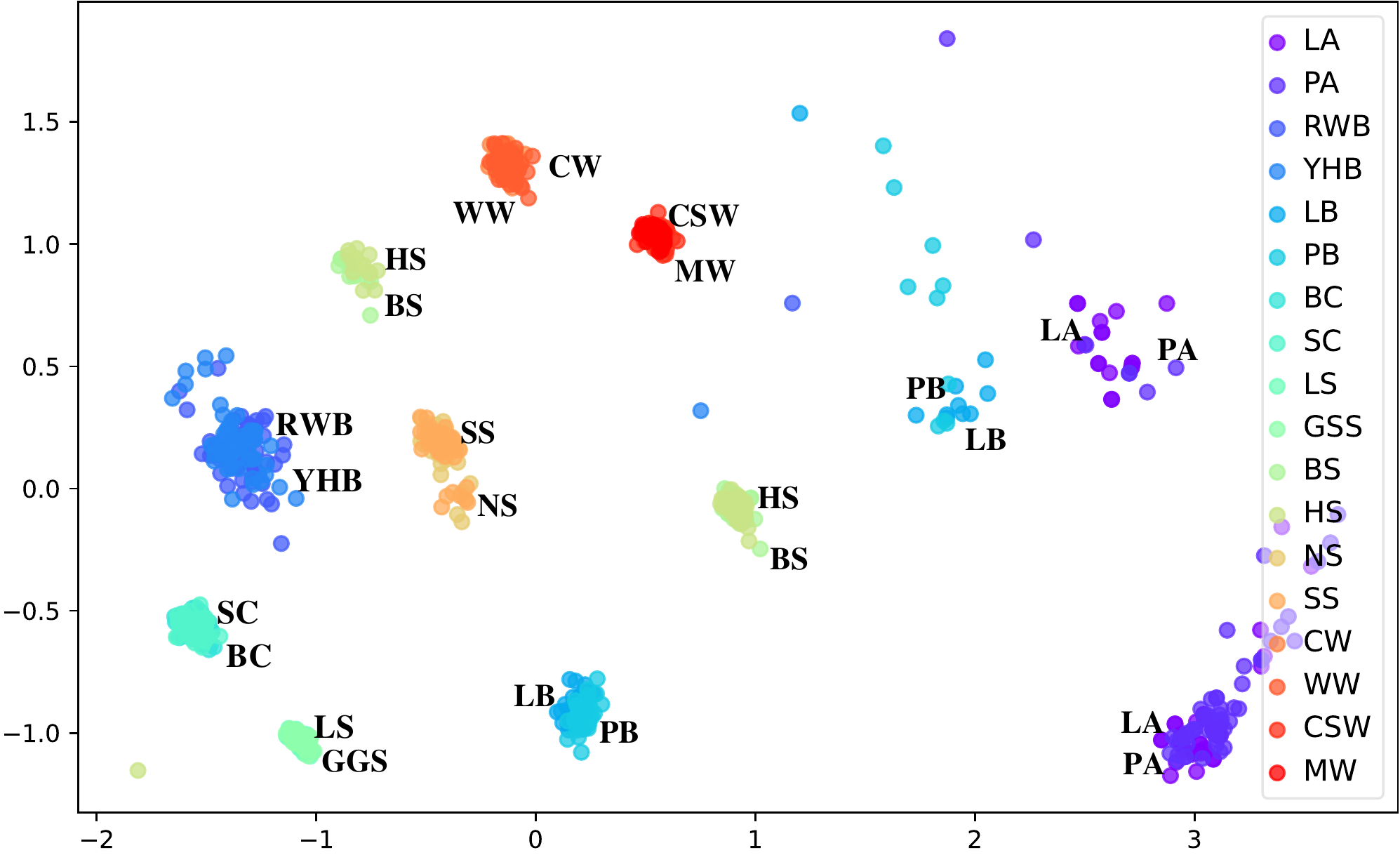}}
\vskip -0.0cm
\caption{TSDA's learned encoding $\e$ for $18$ domains on \emph{CUB-DT}. Domain ``Least Auklet'' (LA) and domain ``Parakeet Auklet'' (PA) are very different from all others in the domain taxonomy (\figref{fig:cub}), correspondingly their encodings are on the far right side, with most of them in the bottom-right corner. Note that each domain has $2$ clusters associated with the binary classification labels.}
\label{fig:cub_vis}
\end{center}
\vskip -0.5in
\end{figure}

\textbf{\emph{CUB-DT}}. \tabref{tab:cub} shows the accuracy for each of the $9$ target domains on \emph{CUB-DT} as well as the average accuracy for different methods. Compared to Source-Only, existing DA methods including DANN, CDANN, ADDA, MDD, and GRDA fail to achieve performance boost in terms of average accuracy. In terms of individual target domain accuracy, ADDA and GRDA are able to improve upon Source-Only in domain ``Seaside Sparrow'' and domain ``Henslow's Sparrow'', respectively. In contrast, our TSDA significantly improves upon all baselines in terms of average accuracy and achieves the highest accuracy in $6$ out of the $9$ target domains by taking full advantage of the domain taxonomy during domain adaptation. 

\figref{fig:cub_vis} plots TSDA's learned encodings $\e$ on all $18$ domains of \emph{CUB-DT}. We can see that the positions of the learned encodings are consistent with the domain taxonomy in~\figref{fig:cub}. For example, domain ``Least Auklet'' (LA) and domain ``Parakeet Auklet'' (PA) share the same parent in the domain taxonomy, and therefore their encodings are close in the figure. Moreover, these two domains are very different from all other domains in the domain taxonomy (\figref{fig:cub}); correspondingly their encodings are on the far right side of the figure, with most of them in the bottom-right corner. 

\section{Conclusion and Future Work}
We propose to characterize domain similarity using a \emph{domain taxonomy}, identify the problem of adaptation across taxonomy-structured domains, and develop taxonomy-structured domain adaptation (TSDA) as the first general DA method to address this problem. We provide theoretical analysis showing that our TSDA retains typical DA methods' capability of uniform alignment when the domain taxonomy is non-informative, and balances domain similarity and domain invariance for other domain taxonomies. As a limitation, our TSDA still assumes the availability of a taxonomy structure to describe relationship among different domains. Therefore from the methodological perspective, it would be interesting future work to explore jointly inferring the domain taxonomy and performing domain adaptation through either conditional or causal approaches~\cite{ICL} while also accounting for the uncertainty of the inferred taxonomy~\cite{TFUncertainty}. From the empirical (application) perspective, it would also be interesting future work to explore taxonomy-structured multi-domain data from other modalities including images~\cite{ReverseAttack}, speech signals~\cite{DGP}, time series~\cite{PDPredict}, wireless signals~\cite{MSA}, etc.

\section*{Acknowledgement}
We would like to thank Bernie Wang and the reviewers/AC for the constructive comments to improve the paper. ZX and
HW are partially supported by NSF Grant IIS-2127918 and an Amazon Faculty Research Award.

\bibliography{example_paper}
\bibliographystyle{icml2023}

%%%%%%%%%%%%%%%%%%%%%%%%%%%%%%%%%%%%%%%%%%%%%%%%%%%%%%%%%%%%%%%%%%%%%%%%%%%%%%%
%%%%%%%%%%%%%%%%%%%%%%%%%%%%%%%%%%%%%%%%%%%%%%%%%%%%%%%%%%%%%%%%%%%%%%%%%%%%%%%
% APPENDIX
%%%%%%%%%%%%%%%%%%%%%%%%%%%%%%%%%%%%%%%%%%%%%%%%%%%%%%%%%%%%%%%%%%%%%%%%%%%%%%%
%%%%%%%%%%%%%%%%%%%%%%%%%%%%%%%%%%%%%%%%%%%%%%%%%%%%%%%%%%%%%%%%%%%%%%%%%%%%%%%
\newpage
\appendix
\onecolumn

\section{Proof}

% Corollary 4.1 ok
% Lemma 4.1 - to be done
% Theorem 4.3 - to be done
\begin{customThm}{4.1}[\textbf{Incompatibility}]%\label{thm:txn_value}
If $\e \indep u$ and $\ell_2$ distance is used in $L_t(\cdot)$, 
\begin{align*}
    L_t(T,E) = 0 \implies \A_{ij} = a, \forall i \neq j \in \gU
\end{align*}
for some $a \in \sZ_{>0}$.
\end{customThm}

\begin{proof}
\begingroup\makeatletter\def\f@size{9}\check@mathfonts
We first expand the loss as
\begin{align*}
     L_t(T,E) = & \frac{1}{2}\int\!\!\!\!\int \sum_{i \neq j} (\A_{ij} \!-\! T(\e, \e'))^2 p(\e, i) p(\e', j)\ d\e \, d\e'\\
              = & \frac{1}{2N^2} \int\!\!\!\!\int \sum_{i \neq j} (\A_{ij} - T(\e, \e'))^2 p(\e) p(\e') \ d\e \, d\e' \\
              = & \frac{1}{2N^2} \sum_{i \neq j} \A_{ij}^2 -2\A_{ij}\mathbb{E}[T(\e, \e')] + \mathbb{E}[T(\e, \e')^2], 
\end{align*}
where the second equality is due to $p(\e, i) = p(\e) p(i) = p(\e) / N$. We also have the equality
\begin{align*}
     \mathbb{E}[T(\e, \e')^2] = \mathbb{E}[T(\e, \e')]^2 + \text{Var}[T(\e_i, \e_j)].
\end{align*}
Combining the two {equalities}, we get

\begin{align*}
L_t(T,E) = & \alpha \text{Var}[T(\e, \e')] + \beta \sum_{i \neq j} [\A_{ij} - \mathbb{E}[T(\e, \e')]]^2 
\geq 0,
\end{align*}
where $\alpha = (N)(N-1)/(2N^2)$ and $\beta = 1/(2N^2)$. Finally,
\begin{align*}
L_t(T,E) = 0 \implies \A_{ij} =  \mathbb{E}[T(\e, \e')], \forall i \neq j,
\end{align*}
completing the proof. 
\endgroup
\end{proof}

\begin{customCor}{4.1}[\textbf{TSDA Generalizes DANN}] %\label{app_cor:tda2dann}
Omitting the predictor, if the taxonomy is non-informative, then the optimum of TSDA is achieved if and only if the embedding distributions of all the domains are the same, i.e. $p(\e|u=1)=\cdots=p(\e|u=N)=p(\e), \forall \e$.
\end{customCor}
\begin{proof}

Based on Theorem 4.1, we know that given the taxonomy is non-informative, we always have $T(\e, \e') = a = A_{ij}$, such that $ L_t(T,E) = 0$ for any $\e, \e'$, (at this time, $\text{Var}[T(\e, \e')]=0$, $\sum_{i \neq j} [\A_{ij} - \mathbb{E}[T(\e, \e')]]^2 = 0$). Thus, to ensure that TSDA is optimal, we only need to ensure that the discriminator achieves its optimum, because $\forall \e$, the taxonomist will always be optimal. The discriminator achieves its optimum if and only $p(\e|u=1)=\cdots=p(\e|u=N)=p(\e), \forall \e$, completing our proof.

\end{proof}

% \blue{\\Guangyuan added the content below.} 
\begin{customThm}{4.2}[\textbf{Uniform Alignment, $\lambda_t$, and $\lambda_d$}]% \label{appthm:lambda_ua}
If $\lambda_t > \lambda_d$ and the
domain taxonomy is not non-informative, 
% and $H(\A)\triangleq \frac{1}{N^2}\sum_{u_1,u_2}\A_{u_1,u_2}>0$, 
% and $H(\A)= H(A_e)>0$
% where $A_e \sim p(A_e)=\frac{1}{N^2-N} \sum_{i \neq j} p(A_e|i, j)$ where $p(A_e|i, j)=\delta(A_e-A_{i, j})$
$\min_{E, T} \max_D -\lambda_d L_d(D,E)+\lambda_t L_t(T,E)$ will not yield uniform alignment 
% (\defref{def:ua})
. \end{customThm}

\begin{proof}
\begingroup\makeatletter\def\f@size{9}\check@mathfonts
Define $A_e \sim p(A_e)=\frac{1}{N^2-N} \sum_{i \neq j} p(A_e|i, j)$ where $p(A_e|i, j)=\delta(A_e-\A_{i, j})$. We model $q(A_e|\e, \e')=\NM(T(\e, \e'), 1)$, which is trainable. Let $L_t(T,E)=-\EB_{p(A_e, \e, \e')}[\log q(A_e|\e, \e')]$, which is another form of $L_t(T,E)$ in the main paper.
\begin{align*}
     & \min_{E} \min_{T} \max_D -\lambda_d L_d(D,E)+\lambda_t [L_t(T,E)]
     \\= & \min_{E} \lambda_d [I(u; \e)-H[u])-\lambda_t [I(A_e; \e, \e')-H(A_e)] 
     \\ =& \lambda_d \{[H(A_e)-H(u)]+  \min_{E} [I(u; \e)-\lambda I(A_e; \e, \e')]\}
\end{align*}
where $\lambda= \frac{\lambda_t}{\lambda_d}$.

The domain taxonomy is not non-informative, and thus $H(A_e)>0$. Define $I^{\#}$ as $I^{\#}(A_e; \e, \e')=H(A_e)=H(\e, \e')$ holds. When $I^{\#}(A_e; \e, \e')=H(A_e)=H(\e, \e')$, we have
\begin{align*}
     \min_{E} & I(u; \e)-\lambda I(A_e; \e, \e')
     \\ & \leq I^{\#}(u; \e)-\lambda I^{\#}(A_e; \e, \e')
 \\ &\leq I^{\#}(u; \e, \e')-\lambda H(A_e)
     \\ &\leq H(\e, \e')-\lambda H(A_e)
     \\& = H(A_e)-\lambda H(A_e)
\end{align*}
     % \\ = I^{\#}(\e;u ; A)+I^{\#}(\e; u|A)-\lambda H(A_e)
When $H(A_e)-\lambda H(A_e) < 0$, i.e., $\lambda=\frac{\lambda_t}{\lambda_d} > 1$, $\min_{E} I(u; \e)-\lambda I(A_e; \e, \e') < 0$. Then, we have
% $I(A_e|\e, \e')> \frac{1}{\lambda}I(u; \e)$ and 
$I(u; \e)>0$ since, if $I(u; \e)=0$, $I(u; \e)-\lambda I(A_e; \e, \e')=0$ must hold. $I(u; \e)>0$ indicates $\min_{E, T} \max_D -\lambda_d L_d(D,E)+\lambda_t L_t(T,E)$ will not lead to uniform alignment.

\endgroup
\end{proof}

Next, we will prove the lemma and the theorem in Sec 4.3. For reference, we restate the minimax game of the alternative method as follows: 
\begin{align}
    \min_E \max_{D_{ij}} \mathbb{E}[\sum_{j \neq i} w_{ij} \log D_{ij} (E(\x)) ]. %\label{eq:pair}
\end{align}

\begin{customLemma}{4.1}[\textbf{Optimal Discriminator}] %\label{lem:opt_dis_dannafdsadfsa}
For every $E$, the optimal $D_{ij}$ of Eq.~(\ref{eq:pair}) satisfies
\begin{align*}
    D_{ij}(\e) =  \frac{p(\e|u=i)}{p(\e|u=i) +p(\e|u=j)}, \forall \e \in \gZ.
\end{align*}
\end{customLemma}

\begin{proof}
For the fixed $E$, \eqnref{eq:pair} could be written as:
\begin{align}
     \mathbb{E}[\sum_{j \neq i} w_{ij} \log D_{ij} (E(\x)) ] &=  \mathbb{E}_{\x,i \sim p(\x,u)}[\sum_{j \neq i} w_{ij} \log D_{ij} (E(\x)) ]\\
     % \mathbb{E}_{x,i \sim p(x,u)}[ \sum_{j=1, j \neq i}^N w_{ij} \log D_{ij}(E(x)) ] \\
    % &\leq \mathbb{E}_{x,d \sim p(x,u)}[\int   \ dx]
    &= \mathbb{E}_{\e,i \sim p(\e,u)}[\sum_{j \neq i} w_{ij} \log D_{ij} (\e) ] \\ 
    % &= \mathbb{E}_{d\sim p(u)} \mathbb{E}_{e\sim p(e|d)} [ \sum_{j=1}^N w_{ij} \log D_{ij}(e)]
    % 展开分为两步，分别是de的积分和枚举所有的i(实质上是离散的变量积分！)!!!!
    &= \frac{1}{2} \int \sum_{i \neq j} w_{ij} (p(\e,i) \log D_{ij}(\e) + p(\e, j) \log (1 - D_{ij}(\e)) )\, d\e \\
    &= \frac{1}{2} \int \sum_{i \neq j} w_{ij} (p(\e|u=i) p(u=i) \log D_{ij}(\e) + p(\e|u=j)p(u=j) \log (1 - D_{ij}(\e)))\, d\e \\
    &= \frac{1}{2} \sum_{i \neq j}  w_{ij} \int   (p(\e|u=i) p(u=i) \log D_{ij}(\e) + p(\e|j)p(u=j) \log (1 - D_{ij}(\e)))\, d\e  \\
    &= \frac{1}{2N} \sum_{i \neq j}  w_{ij} \int   (p(\e|u=i) \log D_{ij}(\e) + p(\e|u=j)\log (1 - D_{ij}(\e)))\, d\e, \label{eq:obj_ij}
\end{align}
where \eqnref{eq:obj_ij} holds because we assume each domain has the same amount of data, and thus for any domain identity $i$, we have $p(u=i) = \frac{1}{N}$.

For function $f({\alpha}) = a \log \alpha + b \log (1 - \alpha)$ with $a,b \in \mathbb{R}^+$, we have $\argmax_{\alpha} f({\alpha}) = \frac{a}{a + b}$. Therefore, to maximize the value function (\eqnref{eq:pair}), we have the optimal $D_{ij}(e)$ as $\frac{p(\e|u=i)}{p(\e|u=i) +p(\e|u=j)}$ for any domain pair $(i,j)$.

%   To maximize $\argmax_{\alpha}~a \log \alpha + b \log (1 - \alpha)=\frac{a}{a + b}$. Therefore the optimal $D_{ij}(e)$ is $  \frac{p(e|i)p(u=i)}{p(e|i)p(i) +p(e|j)p(u=j)}$ ($a = p(e|i)p(i)$, $b = p(e|j)p(j)$). The optimum can be achieved for each term $\int  w_{ij} (p(e|i) p(i) \log D_{ij}(e) + p(e|j)p(j) \log (1 - D_{ij}(e)))\, de$ in~\eqnref{eq:obj_ij} with different $(i,j)$. If $p(i) = p(j)$ holds for every $i,j$ (i.e., each domain has the same number of data points), we have $D_{ij}(E(x)) =  \frac{p(e|i)}{p(e|i) +p(e|j)}$.
\end{proof}

\begin{customThm}{4.3}[\textbf{{Optimal Encoder}}] % \label{thm:opt_value_tsda}
%Given an adjacency matrix $\A$, 
The min-max game in \eqnref{eq:pair} has a tight lower bound:
\begin{align*}
% L_{d'}(D, E) \geq \frac{\log 2}{N} \sum_{i\neq j} w_{ij},
\max_{D_{ij}} \mathbb{E}[\sum_{j \neq i} w_{ij} \log D_{ij} (E(\x)) ]  \geq \frac{\log 2}{N} \sum_{i\neq j} w_{ij},
\end{align*}
where $N$ denotes the number of domains. Furthermore, the equality, i.e., the optimum, is achieved when
\begin{align*}
 p(\e|u=i) = p(\e|u=j),\mbox{ for any } i,j,
\end{align*}
or equivalently, $p(\e|u=i) = p(\e)$.
\end{customThm}

\begin{proof}
Given the optimal discriminators $\{D_{ij}^*\}_{i\neq j}$ based on Lemma 4.1, the value for \eqref{eq:pair} w.r.t. the encoder $E$ is:
% Given the optimal discriminators $D^*= \{D_{ij}^*\}_{i\neq j}$, the loss for the game w.r.t. the encoder $E$ is

\begingroup\makeatletter\def\f@size{9}\check@mathfonts
\begin{align}
    % & L_{D^*}(E) = L_{d'}(D^*,E) \\
    \max_{D_{ij}} \mathbb{E}[\sum_{j \neq i} w_{ij} \log D_{ij} (E(\x)) ]
    &= \sum_{i \neq j}\frac{w_{ij}}{2} \left( p(u=i) \EB_{\e|u=i} [\log D_{ij}^*(\e)] + p(u=j)\EB_{\e|u=j} [\log (1 - D_{ij}^*(\e))] \right)\\
    &= \sum_{i \neq j} \frac{w_{ij}}{2N} \left( \EB_{\e|u=i} [\log D_{ij}^*(\e)] + \EB_{\e|u=j} [\log (1 - D_{ij}^*(\e))] \right. \label{eq:one_over_n}\\
    &= \sum_{i \neq j} \frac{w_{ij}}{2N} \left( \EB_{\e|u=i} \log \frac{p(\e|u=i)}{p(\e|u=i) +p(\e|u=j)} + \EB_{\e|u=j} \log \frac{p(\e|u=j)}{p(\e|u=i) +p(\e|u=j)} \right) \\  
    &= \frac{1}{2N} \sum_{i \neq j} w_{ij}[- 2\log 2 + KL(p(\e|u=i) \Vert \frac{p(\e|u=i) +p(\e|u=j)}{2}) \\
    &\quad \quad+ KL(p(\e|u=j) \Vert \frac{p(\e|u=i) +p(\e|u=j)}{2}) ] \geq \frac{- \log 2}{N} \sum_{i\neq j} w_{ij},\label{eq:gan_bound}
\end{align}
where \eqnref{eq:one_over_n} is due to $p(u=i) = \frac{1}{N}$ for any $i$. The equality  in~\eqnref{eq:gan_bound} holds if and only if $p(\e|u=1) = p(\e|u=2) = \cdots = p(\e|u=N)$. (This can be easily verified using the property $\int p(\e|u=i) d\e = 1$.) Therefore the optimal encoder $E^*$ is achieved if and only if $p(\e|u=1) = p(\e|u=2) = \cdots = p(\e|u=N)$, i.e., $\e \indep u$. 
\endgroup
% $E^* = \argmin_E V_{D^*}(E)$
\end{proof}

\section{Baselines and Implementation}
\label{subsec:implementation}
% We compare TSDA with various state-of-the-art adversarial domain adaptation models, including Domain Adversarial Neural Networks (\textbf{DANN}) \cite{DANN}, Adversarial Discriminative Domain Adaptation (\textbf{ADDA}) \cite{ADDA}, Conditional Domain Adaptation Neural Networks (\textbf{CDANN}) \cite{CDANN}, Margin Disparity Discrepancy (\textbf{MDD}) \cite{MDD} and Graph-Relational Domain Adaptation (\textbf{GRDA}) \cite{GRDA}. We also include results when one trains the model in the source domains is directly test it in the target domains (\textbf{Source-Only} or \textbf{SO} in short). 
% Since each domain in \emph{ImageNet-Attribute-DT} \cite{imagenet_attribute} and \emph{CUB-DT} \cite{cub200} represents a real-world category or species, we report both individual accuracy in each target domains and average accuracy over all target domains.
% All models above are implemented in PyTorch. The balancing hyperparameters $\lambda_d$ and $\lambda_t$ range from $0.1$ to $1$ . For fair comparison, the encoder for all the baselines takes as input the data $\x$, the domain index $u$, and the node embedding $\z$. For \emph{ImageNet-Attribute-DT} and \emph{CUB-DT}, data $\x$ are embeddings extracted from pretrained Resnet-18 \cite{resnet}.%\hao{use 'the Supplement' not Appendix}
\subsection{Model Architecture}
% 2 things:
% fix the annotation (i.e, 不要符号重复，以及是否全部写得正确
% 彻底检查全文，看看还有没有需要补充进supplement的地方

For fair comparison, all baselines and TSDA use the same encoder and predictor. The encoder has the following components:
\begin{itemize}
    \item A \textbf{raw data encoder} embeds the data $\x_l$ into intermediate embeddings $\h_l$.
    \item A \textbf{taxonomy encoder} embeds the domain distance matrix $\A$ and the domain index $u_l$ to the domain embeddings $\z_{u_l}$ with 1 fully connected (FC) layer. We use a taxonomy embedding loss $L_g$ to pretrain the taxonomy encoder:
    \begin{align*}
    L_g  &= \mathbb{E}_{u_1,u_2 \sim p(u)}[ l_g(\Vert \z_{u_1}^\top \z_{u_2} \Vert_2, \A_{u_1, u_2}) ],
    %  - \A_{ij} \log \sigma () - (1 - \A_{ij}) \log (1 - \sigma (\z_i^\top\z_j))],
    \end{align*}
        where $u_1$, $u_2$ are two independent domain identities sampled from $p(u)$, and $l_g$ denotes a regression loss (e.g.,$\ell_2$ distance).
    \item A \textbf{joint encoder} then takes as input both $\h_l$ and $\z_l$ and produces the final embeddings $\e_l$ with 2 FC layers. 
\end{itemize}
For toy datasets \emph{DT-14} and \emph{DT-40}, we use 3 FC layers as the raw data encoder, while for the real dataset \emph{ImageNet-Attribute-DT} and \emph{CUB-DT}, we use PyTorch's default pretrained Resnet-18 as the raw data encoder. 

All the predictors of baselines and TSDA contain 3 FC layers, and all the discriminators have 6 FC layers. For GRDA, we treat every pair of domains that share a common grandparent node as connected, construct GRDA's domain graph, and feed it into its discriminator to recover the graph. 

For the structure of the taxonomist, we first use a 6-FC-layer neural network $T^{'}$ to produce a 2-dimensional taxonomy representation $\t_l$ of the data embedding $\e_l$, and then calculate the $\ell_2$ distance of a pair of taxonomy representations $\t_l$. This can be written as:
% \begin{align*}
%     T(E(\x_1,u_1,\A), E(\x_2,u_2,\A)) = \Vert T^{'}(E(\x_1,u_1,\A)) - T^{'}(E(\x_2,u_2,\A))  \Vert_2 =  \Vert \t_{u_1} - \t_{u_2} \Vert_2
% \end{align*}
\begin{align*}
    T(E(\x_1,u_1,\A), E(\x_2,u_2,\A)) = \Vert T^{'}(E(\x_1,u_1,\A)) - T^{'}(E(\x_2,u_2,\A))  \Vert_2 =  \Vert \t_{1} - \t_{2} \Vert_2
\end{align*}

% TSDA, the taxonomist consists of 6 FC layers and produces a 2-dimensional taxonomy embedding.

\subsection{Other Parameters}
%For each module in TSDA,
We have $\lambda_d$, $\lambda_t$ and $\lambda_e$ as the weights that balance the discriminator loss, the taxonomist loss, and the predictor loss. Note that here, $\lambda_e$ is not necessary theoretically, and we include it only for convenience of hyperparameter tuning. Our loss function could be written as $\lambda_e L_f(E,F)-\lambda_d L_d(D,E)+\lambda_t L_t(T,E)$, where $\lambda_d$ and $\lambda_e$ range from 0.1 to 1 and $\lambda_t$ ranges from 0.1 to 10. During the hyperparameter tuning, we always ensure that $\lambda_t > \lambda_d$ (\thmref{thm:lambda_ua}). We use Adam optimizer \cite{Adam} for all models with learning rates from $1 \times 10^{-4}$ to $1 \times 10^{-6}$. The input data and the domain distance matrix $\A$ are normalized according to its mean and variance. All experiments are run on NVDIA GeForce RTX 2080 Ti GPUs.

\tabref{tab:hyper_param_analysis} shows the experiment results of various $\lambda_d$,  $\lambda_t$ combination on DT-14. TSDA achieves robust performance when $\lambda_t > \lambda_d$ ($>90\%$ accuracy), while suffers from performance loss when $\lambda_t = \lambda_d$ ($84.1\%$ accuracy). This is in line with the aforementioned \thmref{thm:lambda_ua}'s conclusion.

\begin{table*}[!thp]
\vskip -0.1cm
\centering
% \begin{scriptsize}
  \caption{Sensitivity of hyper-parameters $\lambda_d$ and $\lambda_t$ on DT-14. $\lambda_e$ is fixed at 0.5. The accuracy of TSDA remains stable as long as $\lambda_t > \lambda_d$.}
  \label{tab:hyper_param_analysis}
  \vskip 0.1cm
  \centering
  \begin{tabular}{l|ccc} \toprule
         & $\lambda_t=1$ & $\lambda_t=2$ & $\lambda_t=4$ \\\midrule
   $\lambda_d=0.25$ & 100.0 & 100.0 & 100.0 \\
   $\lambda_d=0.5$ & 97.9 & 99.9 & 96.5 \\
   $\lambda_d=1$ & 84.1 & 90.6 & 91.7 \\
         
     \bottomrule
  \end{tabular}
% \end{scriptsize}
\vskip -0.5cm
\end{table*}

\subsection{Training Procedure}
We implement the minimax game by alternately training modules of TSDA until convergence in the following two steps:
\begin{enumerate}
    \item We fix the encoder $E$, the taxonomist $T$ and the predictor $F$ and optimize the discriminator $D$. With encoding generated from $E$, we use the disciminator loss $L_d(D,E)$ to train the discriminator.
    \item We fix the discriminator $D$ and minimize $\lambda_e L_f(E,F)-\lambda_d L_d(D,E)+\lambda_t L_t(T,E)$ to train the encoder $E$, the taxonomist $T$ and, the predictor $F$.
\end{enumerate}

We summarize the training procedure formally in Algorithm~\ref{alg:training}.
\begin{algorithm}
\caption{TSDA Training}\label{alg:training}
\begin{algorithmic}[1]
\STATE Specify the distance matrix $\A$
\STATE Initialize the encoder $E$, the taxonomist $T$, the predictor $F$ and the discriminator $D$ networks.
\STATE Initialize the $D$ optimizer and the $ETF$ optimizer for $E$, $T$, $F$.
\FOR{each epoch}
  \FOR{each mini-batch of data $\x_l$, domain index $u_l$ with size $m$}
    \STATE Calculate the gradient for $D$: $\nabla_D = \nabla_{\theta_D} \frac{1}{m} L_d(D,E)(\x_l,u_l,\A)$.
    \STATE Update the discriminator weights with $D$ optimizer using $\nabla_D$.

    \STATE Calculate the gradient for $E$, $T$, $F$: $\nabla_{E,T, F} = \nabla_{\theta_{E,T,F}} \frac{1}{m} [\lambda_e L_f(E,F)-\lambda_d L_d(D,E)+\lambda_t L_t(T,E)](\x_l,u_l,\A)$.

    \STATE Update the weights of the encoder, the taxonomist and the predictor with $ETF$ optimizer using $\nabla_{E,T, F}$.

  \ENDFOR
\ENDFOR
\end{algorithmic}
\end{algorithm}

\section{Inference Procedure}
The inference procedure of TSDA is formally presented in Algorithm~\ref{alg:inference}. We only need the encoder $E$ and the predictor $F$ during inference to perform prediction.

\begin{algorithm}
\caption{TSDA Inference}\label{alg:inference}
\begin{algorithmic}[1]
\STATE Load the encoder $E$, the predictor $F$.
\STATE Load the testing data $X_{test}$.
\FOR{each example $x_l$ in $X_{test}$}
    \STATE Predict the output $\hat{y_l}$ = $F(E(x_l))$.
    \STATE Store the prediction $\hat{y_l}$.
\ENDFOR
\STATE Evaluate the performance of the model using desired metric(s) on the predicted ($\hat{y_l}$) and true labels ($y_l$).
\end{algorithmic}
\end{algorithm}

\section{Ablation Study}
We perform an ablation study to demonstrate the effectiveness of the two key components, the discriminator and the taxonomist, in TSDA. The results in~\tabref{tab:ablation} show that without either component, we can observe significant performance drops in most tasks. TSDA without the discriminator is no longer an adversarial domain adaptation framework, which shows that simply aligning similar domains together does not perform well. TSDA without the taxonomist is equivalent to the baseline DANN, and its results reveal that without necessary taxonomy information, performance suffers. The ablation study illustrates that all components contribute to the full model TSDA. 
\begin{table*}[!thp]
\vskip -0.1cm
\centering
% \begin{scriptsize}
  \caption{Accuracy of TSDA compared with TSDA without discriminator or taxonomist on all four datasets as well as the average for all tasks. Note that TSDA without Taxonomist is in fact DANN. We mark the best average accuracy with \textbf{bold face}}
  \label{tab:ablation}
  \vskip 0.1cm
  \centering
  \begin{tabular}{lcccccccc} \toprule
   Target & DT-14 & DT-40 & ImageNet-Attribute-DT & CUB-DT & Average\\\midrule
   TSDA \emph{w/o Discriminator} & 34.5 & 42.7 & 62.0 & 80.6 & 55.0\\
   TSDA \emph{w/o Taxonomist} & 68.8 & 55.4 & 75.3 & 66.9 & 66.6\\
   TSDA & \textbf{100.0} & \textbf{82.6} & \textbf{80.7} & \textbf{82.4} & \textbf{86.4}\\
         
     \bottomrule
  \end{tabular}
% \end{scriptsize}
\vskip -0.5cm
\end{table*}

\section{Larger Figures}
In this section, we provide larger versions of figures for TSDA's learned encoding visualization in the main paper. Results are shown in \figref{fig:im2} and \figref{fig:im3}.

\begin{figure*}[!t]
% \vskip -0.8cm
\begin{center}
% \centerline{\includegraphics[width=\columnwidth *2]{pic/expr1.pdf}}
\centerline{\includegraphics[width=0.99\textwidth]{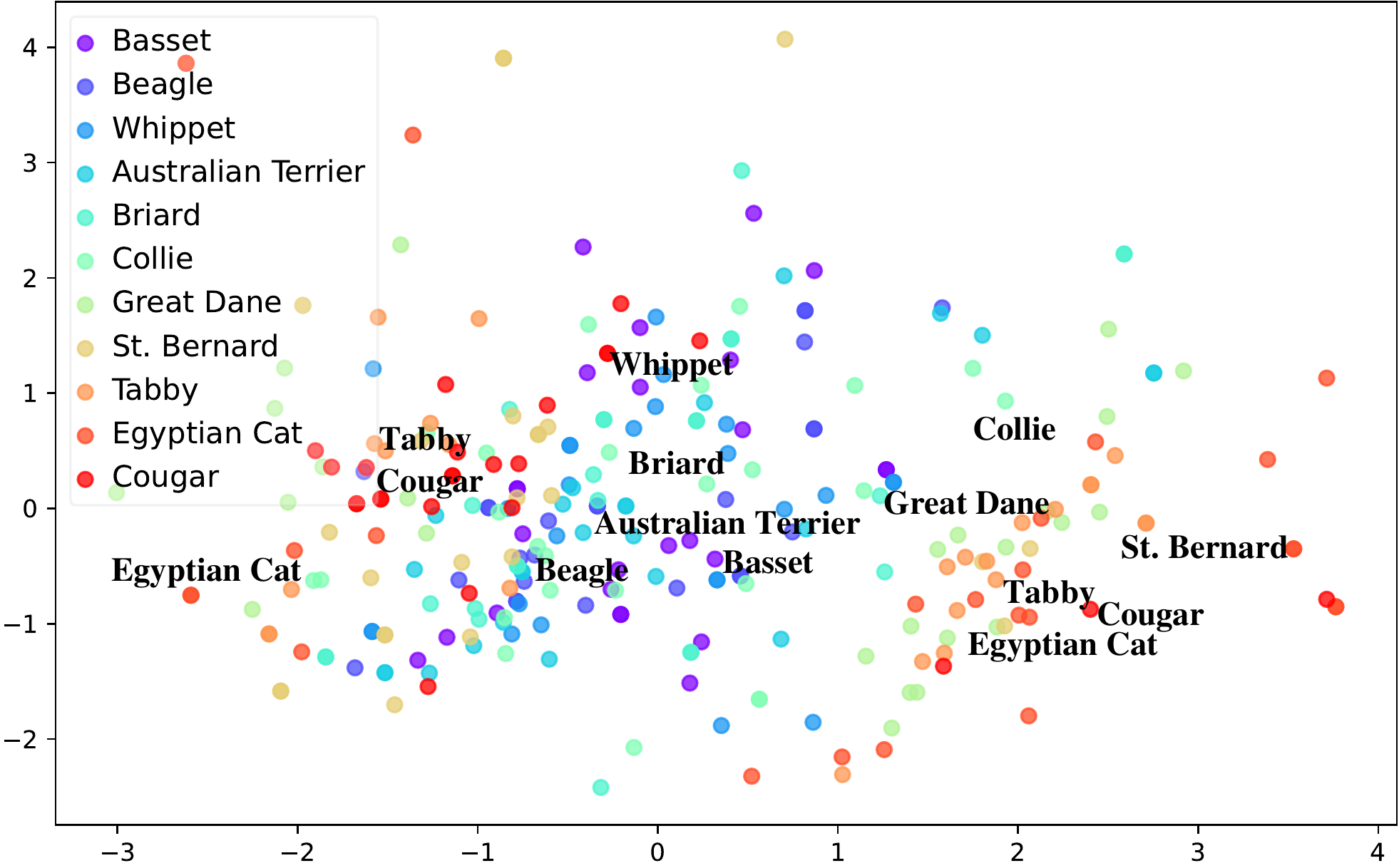}}
\caption{TSDA's learned encoding $\e$ for $11$ domains on \emph{ImageNet-Attribute-DT}.}
\label{fig:im2}
\end{center}
\vskip -0.3in
\end{figure*}

\begin{figure*}[!t]
% \vskip -0.8cm
\begin{center}
% \centerline{\includegraphics[width=\columnwidth *2]{pic/expr1.pdf}}
\centerline{\includegraphics[width=0.99\textwidth]{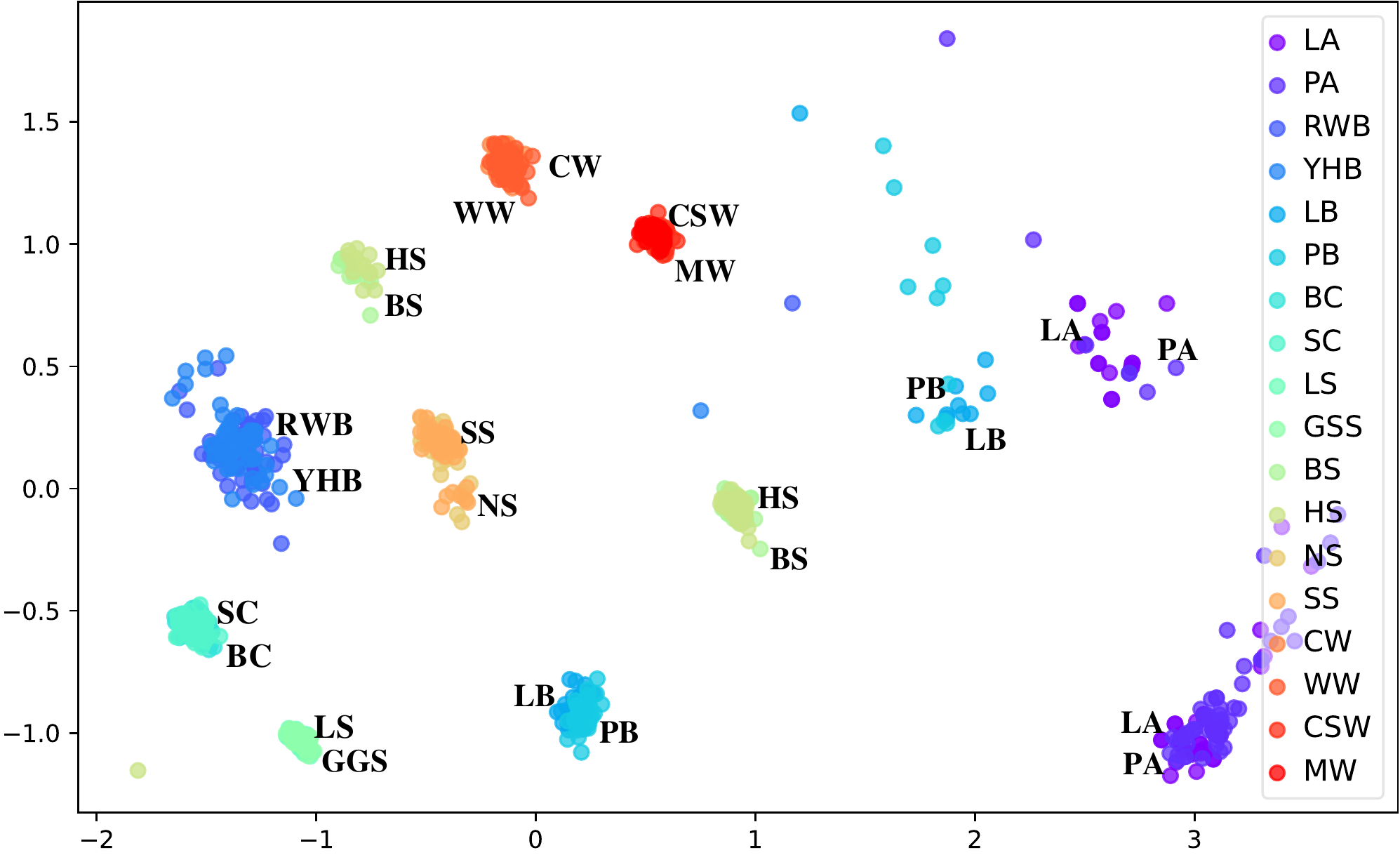}}
\caption{TSDA's learned encoding $\e$ for $18$ domains on \emph{CUB-DT}.}
\label{fig:im3}
\end{center}
\vskip -0.3in
\end{figure*}

% \section{Hyper-parameter Analysis}

\section{Additional Discussions}
\textbf{Adversarial Training versus Non-Adversarial Methods for Domain Alignment.}
The number of new adversarial training methods in recent literature has gradually reduced. However, to the best of our knowledge, adversarial training is still the state of the art for domain alignment. While it is true that non-adversarial approaches have been proposed and shown to be effective, such as maximum mean discrepancy (MMD) and entropy minimization \cite{entropy_min}, adversarial training still achieves state-of-the-art performance in many cases \cite{CIDA, GRDA}; for example, \citet{GRDA} (one of our baseline) has shown that its adversarial DA method could outperform state-of-the-art non-adversarial methods such as \citet{entropy_min}.
Additionally, since our TSDA's taxonomist is cooperative rather than adversarial, it can potentially be incorporated into non-adversarial methods to improve their performance in our new taxonomy-structured DA setting too, which would be interesting future work.

\textbf{Not All Datasets Have Domain Taxonomies as Additional Information. }
While taxonomy is not readily available in every dataset, we believe that such structure naturally formalize many nested, hierarchical domains in the real world; typical examples include product taxonomies in e-commerce and disease taxonomies in healthcare. We therefore see our work as a pilot study to demonstrate the benefit of leveraging domain taxonomies during domain adaptation; with this, we hope to encourage the community to pay more attention to such taxonomy structure both during methodology development and during data collection.

On the other hand, we might also conduct automatic taxonomy induction in an unsupervised way. For example, \citet{domain2vec} and \citet{VDI} show that it is possible to learn domain embeddings or domain indices that capture the relations between different domains in an unsupervised manner. Therefore, we can use \citet{domain2vec} and \citet{VDI} to first infer the domain embeddings or indices, construct a domain taxonomy according to these domain embeddings or indices, and then apply our TSDA. This would be interesting future work.

\textbf{Comparsion with \citet{das2018graph, yang2019cross, pilanci2020domain}.}
All these works use graph matching as a domain discrepancy metric. \citet{das2018graph} and \citet{yang2019cross}  propose to align the ``source graph" and ``target graph" constructed from source and target domains, respectively, where \emph{each data point is a node in the graph}. On the other hand, \citet{pilanci2020domain} aims to perform domain adaptation in a setting where \emph{each data point is a graph}. In contrast, our TSDA focuses on the setting where each domain is a node in the domain taxonomy. Additionally, \citet{das2018graph, yang2019cross, pilanci2020domain} focus on the setting with a single source domain and a single target domain, while our TSDA focuses on multiple source domains and target domains. 
Therefore, \citet{das2018graph, yang2019cross, pilanci2020domain} are \emph{not applicable} in our setting.

\textbf{Extension to Source-Free Setting.} 
Our method can be naturally extended to the source-free setting. Specifically, we could apply our approach only to target domains, and use the pretrained model to regularize the training process. We can keep the classifier $C$ fixed and train the encoder $E$, taxonomist $T$, and discriminator $D$ given the target domain taxonomy. Meanwhile, we encourage the model to produce similar predictions as the pretrained model. 

Formally, denote the input as $x$, the pretrained encoder as $E'$, and the encoder after adaptation as $E$. In this case, we can use DIST$( C(E'(x)) , C(E(x)) )$ as an additional regularization term to regularize the training process. Here ``DIST" refers to the distance between two predictions, which could be cross-entropy (for classification tasks) or L2 (for regression tasks). The final objective function will then become the sum of DIST$( C(E'(x)) , C(E(x)) )$ and \eqnref{eq:min_max_game}.

\begin{table*}[!thp]
\vskip -0.1cm
\centering
% \begin{scriptsize}
  \caption{ Accuracy on DT-14 with different numbers of samples.}
  \label{tab:sample_complexity}
  \vskip 0.1cm
  \centering
  \begin{tabular}{ccc} \toprule
         Sample Number  & DANN   & TSDA   \\\midrule
            50\%            & 61.6   & 99.2   \\
          75\%            & 65.6   & 99.9   \\
          100\%           & 68.8   & 100.0  \\
     \bottomrule
  \end{tabular}
% \end{scriptsize}
\vskip -0.5cm
\end{table*}

\textbf{Sample Complexity.} 
From an additional experiment on DT-14 (\tabref{tab:sample_complexity}), we can see that as the number of samples in the source domain decreases (e.g., only 50\% of the original samples), the baseline method (DANN) suffers from a performance drop. In contrast, TSDA remains stable and outperforms the baseline by a significant margin.

\textbf{Limitation of Our Method.} 
 As mentioned in Corollary 4.1, our method will degenerate into {DANN \cite{DANN}} when the taxonomy is non-informative, i.e., when the distance between every pair of domains is identical (e.g., a flat taxonomy). In this case, our method essentially reduces to the standard DANN model. 
Another limitation is that the domain taxonomy should provide a suitable inductive bias to the learning task (the domains are similar when they are closer in the taxonomy). Taxonomies without such inductive bias yield no benefit, or even do harm to the domain adaptation performance.

%%%%%%%%%%%%%%%%%%%%%%%%%%%%%%%%%%%%%%%%%%%%%%%%%%%%%%%%%%%%%%%%%%%%%%%%%%%%%%%
%%%%%%%%%%%%%%%%%%%%%%%%%%%%%%%%%%%%%%%%%%%%%%%%%%%%%%%%%%%%%%%%%%%%%%%%%%%%%%%

\end{document}

%% file: defs.tex
\def\A{{\bf A}}

\def\e{{\bf e}}

\def\h{{\bf h}}

\def\I{{\bf I}}

\def\t{{\bf t}}

\def\x{{\bf x}}

\def\z{{\bf z}}

\def\0{{\bf 0}}
\def\1{{\bf 1}}

\def\NM{{\mathcal N}}

\def\EB{{\mathbb E}}

\def\muu{\mbox{\boldmath$\mu$\unboldmath}}

\newcommand{\indep}{{\;\bot\!\!\!\!\!\!\bot\;}}
\def\argmax{\mathop{\rm argmax}}

\newtheorem{theorem}{Theorem}
\newtheorem{lemma}{Lemma}
\newtheorem{definition}{Definition}

\newtheorem{cor}{Corollary}
\numberwithin{theorem}{section}
\numberwithin{lemma}{section}
\numberwithin{remark}{section}
\numberwithin{cor}{section}
\numberwithin{proposition}{section}
\numberwithin{definition}{section}

\newtheorem{innercustomgeneric}{\customgenericname}
\providecommand{\customgenericname}{}
\newcommand{\newcustomtheorem}[2]{%
  \newenvironment{#1}[1]
  {%
   \renewcommand\customgenericname{#2}%
   \renewcommand\theinnercustomgeneric{##1}%
   \innercustomgeneric
  }
  {\endinnercustomgeneric}
}

\newcustomtheorem{customThm}{Theorem}
\newcustomtheorem{customLemma}{Lemma}
\newcustomtheorem{customCor}{Corollary}
\newcustomtheorem{customProposition}{Proposition}

\newcommand{\tabref}[1]{Table~\ref{#1}}
\newcommand{\secref}[1]{Sec.~\ref{#1}}
\newcommand{\figref}[1]{Fig.~\ref{#1}}

\newcommand{\thmref}[1]{Theorem~\ref{#1}}

\newcommand{\defref}[1]{Definition~\ref{#1}}

\newcommand{\crlref}[1]{Corollary~\ref{#1}}
\newcommand{\eqnref}[1]{Eq.~(\ref{#1})}

%% file: math_commands.tex
\usepackage{amsmath,amsfonts,bm}

\def\1{\bm{1}}

\def\mA{{\bf A}}

\DeclareMathAlphabet{\mathsfit}{\encodingdefault}{\sfdefault}{m}{sl}
\SetMathAlphabet{\mathsfit}{bold}{\encodingdefault}{\sfdefault}{bx}{n}

\def\gT{{\mathcal{T}}}
\def\gU{{\mathcal{U}}}

\def\gX{{\mathcal{X}}}
\def\gY{{\mathcal{Y}}}
\def\gZ{{\mathcal{Z}}}

\def\sZ{{\mathbb{Z}}}

%% file: example_paper.bbl
\begin{thebibliography}{60}
\providecommand{\natexlab}[1]{#1}
\providecommand{\url}[1]{\texttt{#1}}
\expandafter\ifx\csname urlstyle\endcsname\relax
  \providecommand{\doi}[1]{doi: #1}\else
  \providecommand{\doi}{doi: \begingroup \urlstyle{rm}\Url}\fi

\bibitem[Ben-David et~al.(2010)Ben-David, Blitzer, Crammer, Kulesza, Pereira,
  and Vaughan]{ben2010theory}
Ben-David, S., Blitzer, J., Crammer, K., Kulesza, A., Pereira, F., and Vaughan,
  J.~W.
\newblock A theory of learning from different domains.
\newblock \emph{Machine learning}, 79\penalty0 (1):\penalty0 151--175, 2010.

\bibitem[Chen et~al.(2019)Chen, Zhuang, Liang, and Lin]{blending}
Chen, Z., Zhuang, J., Liang, X., and Lin, L.
\newblock Blending-target domain adaptation by adversarial meta-adaptation
  networks.
\newblock In \emph{Proceedings of the IEEE/CVF Conference on Computer Vision
  and Pattern Recognition}, pp.\  2248--2257, 2019.

\bibitem[Dai et~al.(2019)Dai, Sohn, Tsai, Carin, and Chandraker]{domain_bridge}
Dai, S., Sohn, K., Tsai, Y.-H., Carin, L., and Chandraker, M.
\newblock Adaptation across extreme variations using unlabeled domain bridges.
\newblock \emph{arXiv preprint arXiv:1906.02238}, 2019.

\bibitem[Das \& Lee(2018)Das and Lee]{das2018graph}
Das, D. and Lee, C.~G.
\newblock Graph matching and pseudo-label guided deep unsupervised domain
  adaptation.
\newblock In \emph{Artificial Neural Networks and Machine Learning--ICANN 2018:
  27th International Conference on Artificial Neural Networks, Rhodes, Greece,
  October 4-7, 2018, Proceedings, Part III 27}, pp.\  342--352. Springer, 2018.

\bibitem[Ding et~al.(2022)Ding, Ma, Deoras, Wang, and Wang]{ZESRec}
Ding, H., Ma, Y., Deoras, A., Wang, Y., and Wang, H.
\newblock Zero-shot recommender systems.
\newblock In \emph{ICLR Workshop on Deep Generative Models for Highly
  Structured Data}, 2022.

\bibitem[Ganin et~al.(2016)Ganin, Ustinova, Ajakan, Germain, Larochelle,
  Laviolette, Marchand, and Lempitsky]{DANN}
Ganin, Y., Ustinova, E., Ajakan, H., Germain, P., Larochelle, H., Laviolette,
  F., Marchand, M., and Lempitsky, V.
\newblock Domain-adversarial training of neural networks.
\newblock \emph{JMLR}, 17\penalty0 (1):\penalty0 2096--2030, 2016.

\bibitem[Gong et~al.(2021)Gong, Danelljan, Dai, Wang, Paudel, Chhatkuli, Yu,
  and Van~Gool]{gong2021tada}
Gong, R., Danelljan, M., Dai, D., Wang, W., Paudel, D.~P., Chhatkuli, A., Yu,
  F., and Van~Gool, L.
\newblock Tada: Taxonomy adaptive domain adaptation.
\newblock \emph{Withdrawn from ICLR 2022}, 2021.

\bibitem[Gong et~al.(2022)Gong, Danelljan, Dai, Paudel, Chhatkuli, Yu, and
  Gool]{TACS}
Gong, R., Danelljan, M., Dai, D., Paudel, D.~P., Chhatkuli, A., Yu, F., and
  Gool, L.~V.
\newblock Tacs: Taxonomy adaptive cross-domain semantic segmentation.
\newblock In \emph{ECCV}, 2022.

\bibitem[Goodfellow et~al.(2014)Goodfellow, Pouget-Abadie, Mirza, Xu,
  Warde-Farley, Ozair, Courville, and Bengio]{GAN}
Goodfellow, I., Pouget-Abadie, J., Mirza, M., Xu, B., Warde-Farley, D., Ozair,
  S., Courville, A., and Bengio, Y.
\newblock Generative adversarial nets.
\newblock In \emph{NIPS}, pp.\  2672--2680, 2014.

\bibitem[Grandvalet \& Bengio(2004)Grandvalet and Bengio]{entropy_min}
Grandvalet, Y. and Bengio, Y.
\newblock Semi-supervised learning by entropy minimization.
\newblock \emph{Advances in neural information processing systems}, 17, 2004.

\bibitem[He \& Peng(2019)He and Peng]{cub200}
He, X. and Peng, Y.
\newblock Fine-grained visual-textual representation learning.
\newblock \emph{IEEE Transactions on Circuits and Systems for Video
  Technology}, 30\penalty0 (2):\penalty0 520--531, 2019.

\bibitem[Huang et~al.(2019)Huang, Wang, and Mak]{RPPU}
Huang, H., Wang, H., and Mak, B.
\newblock Recurrent poisson process unit for speech recognition.
\newblock In \emph{AAAI}, volume~33, pp.\  6538--6545, 2019.

\bibitem[Huang et~al.(2020)Huang, Xue, Wang, and Wang]{DGP}
Huang, H., Xue, F., Wang, H., and Wang, Y.
\newblock Deep graph random process for relational-thinking-based speech
  recognition.
\newblock In \emph{ICML}, 2020.

\bibitem[Jin et~al.(2022)Jin, Park, Maddix, Wang, and Wang]{jin2022domain}
Jin, X., Park, Y., Maddix, D., Wang, H., and Wang, Y.
\newblock Domain adaptation for time series forecasting via attention sharing.
\newblock In \emph{International Conference on Machine Learning}, pp.\
  10280--10297. PMLR, 2022.

\bibitem[Kingma \& Ba(2015)Kingma and Ba]{Adam}
Kingma, D.~P. and Ba, J.
\newblock Adam: {A} method for stochastic optimization.
\newblock In \emph{ICLR}, 2015.

\bibitem[Kumar et~al.(2020)Kumar, Ma, and Liang]{kumar2020understanding}
Kumar, A., Ma, T., and Liang, P.
\newblock Understanding self-training for gradual domain adaptation.
\newblock In \emph{International Conference on Machine Learning}, pp.\
  5468--5479. PMLR, 2020.

\bibitem[Kuroki et~al.(2019)Kuroki, Charoenphakdee, Bao, Honda, Sato, and
  Sugiyama]{UDA-SGD}
Kuroki, S., Charoenphakdee, N., Bao, H., Honda, J., Sato, I., and Sugiyama, M.
\newblock Unsupervised domain adaptation based on source-guided discrepancy.
\newblock In \emph{AAAI}, pp.\  4122--4129, 2019.

\bibitem[Long et~al.(2018)Long, Cao, Wang, and Jordan]{CDAN}
Long, M., Cao, Z., Wang, J., and Jordan, M.~I.
\newblock Conditional adversarial domain adaptation.
\newblock In \emph{NIPS}, pp.\  1647--1657, 2018.

\bibitem[Mancini et~al.(2019)Mancini, Bulo, Caputo, and
  Ricci]{mancini2019adagraph}
Mancini, M., Bulo, S.~R., Caputo, B., and Ricci, E.
\newblock Adagraph: Unifying predictive and continuous domain adaptation
  through graphs.
\newblock In \emph{CVPR}, pp.\  6568--6577, 2019.

\bibitem[Mao et~al.(2021)Mao, Chiquier, Wang, Yang, and
  Vondrick]{ReverseAttack}
Mao, C., Chiquier, M., Wang, H., Yang, J., and Vondrick, C.
\newblock Adversarial attacks are reversible with natural supervision.
\newblock In \emph{ICCV}, 2021.

\bibitem[Maria~Carlucci et~al.(2017)Maria~Carlucci, Porzi, Caputo, Ricci, and
  Rota~Bulo]{maria2017autodial}
Maria~Carlucci, F., Porzi, L., Caputo, B., Ricci, E., and Rota~Bulo, S.
\newblock Autodial: Automatic domain alignment layers.
\newblock In \emph{Proceedings of the IEEE international conference on computer
  vision}, pp.\  5067--5075, 2017.

\bibitem[Mi et~al.(2022)Mi, Wang, Tian, and Shavit]{TFUncertainty}
Mi, L., Wang, H., Tian, Y., and Shavit, N.
\newblock Training-free uncertainty estimation for neural networks.
\newblock In \emph{AAAI}, 2022.

\bibitem[Miller(1995)]{wordnet}
Miller, G.~A.
\newblock Wordnet: a lexical database for english.
\newblock \emph{Communications of the ACM}, 38\penalty0 (11):\penalty0 39--41,
  1995.

\bibitem[Nguyen-Meidine et~al.(2021)Nguyen-Meidine, Belal, Kiran, Dolz,
  Blais-Morin, and Granger]{DA_distillation}
Nguyen-Meidine, L.~T., Belal, A., Kiran, M., Dolz, J., Blais-Morin, L.-A., and
  Granger, E.
\newblock Unsupervised multi-target domain adaptation through knowledge
  distillation.
\newblock In \emph{Proceedings of the IEEE/CVF Winter Conference on
  Applications of Computer Vision}, pp.\  1339--1347, 2021.

\bibitem[Ouyang et~al.(2015)Ouyang, Li, Zeng, and Wang]{imagenet_attribute}
Ouyang, W., Li, H., Zeng, X., and Wang, X.
\newblock Learning deep representation with large-scale attributes.
\newblock In \emph{Proceedings of the IEEE International Conference on Computer
  Vision}, pp.\  1895--1903, 2015.

\bibitem[Pan \& Yang(2009)Pan and Yang]{TLSurvey}
Pan, S.~J. and Yang, Q.
\newblock A survey on transfer learning.
\newblock \emph{TKDE}, 22\penalty0 (10):\penalty0 1345--1359, 2009.

\bibitem[Pan \& Yang(2010)Pan and Yang]{pan2010survey}
Pan, S.~J. and Yang, Q.
\newblock A survey on transfer learning.
\newblock \emph{IEEE Transactions on knowledge and data engineering},
  22\penalty0 (10):\penalty0 1345--1359, 2010.

\bibitem[Pan et~al.(2010)Pan, Tsang, Kwok, and Yang]{MMD}
Pan, S.~J., Tsang, I.~W., Kwok, J.~T., and Yang, Q.
\newblock Domain adaptation via transfer component analysis.
\newblock \emph{TNN}, 22\penalty0 (2):\penalty0 199--210, 2010.

\bibitem[Peng et~al.(2019)Peng, Bai, Xia, Huang, Saenko, and Wang]{moment}
Peng, X., Bai, Q., Xia, X., Huang, Z., Saenko, K., and Wang, B.
\newblock Moment matching for multi-source domain adaptation.
\newblock In \emph{Proceedings of the IEEE/CVF International Conference on
  Computer Vision}, pp.\  1406--1415, 2019.

\bibitem[Peng et~al.(2020)Peng, Li, and Saenko]{domain2vec}
Peng, X., Li, Y., and Saenko, K.
\newblock Domain2vec: Domain embedding for unsupervised domain adaptation.
\newblock \emph{arXiv preprint arXiv:2007.09257}, 2020.

\bibitem[Pilanc{\i} \& Vural(2020)Pilanc{\i} and Vural]{pilanci2020domain}
Pilanc{\i}, M. and Vural, E.
\newblock Domain adaptation on graphs by learning aligned graph bases.
\newblock \emph{IEEE Transactions on Knowledge and Data Engineering},
  34\penalty0 (2):\penalty0 587--600, 2020.

\bibitem[Prabhu et~al.(2021)Prabhu, Khare, Kartik, and
  Hoffman]{prabhu2021sentry}
Prabhu, V., Khare, S., Kartik, D., and Hoffman, J.
\newblock Sentry: Selective entropy optimization via committee consistency for
  unsupervised domain adaptation.
\newblock In \emph{Proceedings of the IEEE/CVF International Conference on
  Computer Vision}, pp.\  8558--8567, 2021.

\bibitem[Rangapuram et~al.(2023)Rangapuram, Kapoor, Nirwan, Mercado,
  Januschowski, Wang, and Bohlke-Schneider]{rangapuram2023coherent}
Rangapuram, S.~S., Kapoor, S., Nirwan, R.~S., Mercado, P., Januschowski, T.,
  Wang, Y., and Bohlke-Schneider, M.
\newblock Coherent probabilistic forecasting of temporal hierarchies.
\newblock In \emph{International Conference on Artificial Intelligence and
  Statistics}, pp.\  9362--9376. PMLR, 2023.

\bibitem[Redko et~al.(2020)Redko, Morvant, Habrard, Sebban, and
  Bennani]{redko2020survey}
Redko, I., Morvant, E., Habrard, A., Sebban, M., and Bennani, Y.
\newblock A survey on domain adaptation theory: learning bounds and theoretical
  guarantees.
\newblock \emph{arXiv preprint arXiv:2004.11829}, 2020.

\bibitem[Saito et~al.(2018)Saito, Watanabe, Ushiku, and Harada]{MCD}
Saito, K., Watanabe, K., Ushiku, Y., and Harada, T.
\newblock Maximum classifier discrepancy for unsupervised domain adaptation.
\newblock In \emph{CVPR}, pp.\  3723--3732, 2018.

\bibitem[Sankaranarayanan et~al.(2018)Sankaranarayanan, Balaji, Castillo, and
  Chellappa]{GTA}
Sankaranarayanan, S., Balaji, Y., Castillo, C.~D., and Chellappa, R.
\newblock Generate to adapt: Aligning domains using generative adversarial
  networks.
\newblock In \emph{CVPR}, pp.\  8503--8512, 2018.

\bibitem[Sun \& Saenko(2016)Sun and Saenko]{CORAL}
Sun, B. and Saenko, K.
\newblock Deep {CORAL:} correlation alignment for deep domain adaptation.
\newblock In \emph{ICCV workshop on Transferring and Adapting Source Knowledge
  in Computer Vision (TASK-CV)}, pp.\  443--450, 2016.

\bibitem[Tasar et~al.(2020)Tasar, Tarabalka, Giros, Alliez, and
  Clerc]{tasar2020standardgan}
Tasar, O., Tarabalka, Y., Giros, A., Alliez, P., and Clerc, S.
\newblock Standardgan: Multi-source domain adaptation for semantic segmentation
  of very high resolution satellite images by data standardization.
\newblock In \emph{Proceedings of the IEEE/CVF Conference on Computer Vision
  and Pattern Recognition Workshops}, pp.\  192--193, 2020.

\bibitem[Tzeng et~al.(2014)Tzeng, Hoffman, Zhang, Saenko, and Darrell]{DDC}
Tzeng, E., Hoffman, J., Zhang, N., Saenko, K., and Darrell, T.
\newblock Deep domain confusion: Maximizing for domain invariance.
\newblock \emph{arXiv preprint arXiv:1412.3474}, 2014.

\bibitem[Tzeng et~al.(2017)Tzeng, Hoffman, Saenko, and Darrell]{ADDA}
Tzeng, E., Hoffman, J., Saenko, K., and Darrell, T.
\newblock Adversarial discriminative domain adaptation.
\newblock In \emph{CVPR}, pp.\  7167--7176, 2017.

\bibitem[Wang(2017)]{BDLThesis}
Wang, H.
\newblock \emph{Bayesian Deep Learning for Integrated Intelligence: Bridging
  the Gap between Perception and Inference}.
\newblock PhD thesis, Hong Kong University of Science and Technology, 2017.

\bibitem[Wang \& Yeung(2016)Wang and Yeung]{BDL}
Wang, H. and Yeung, D.-Y.
\newblock Towards bayesian deep learning: A framework and some existing
  methods.
\newblock \emph{TDKE}, 28\penalty0 (12):\penalty0 3395--3408, 2016.

\bibitem[Wang \& Yeung(2020)Wang and Yeung]{BDLSurvey}
Wang, H. and Yeung, D.-Y.
\newblock A survey on bayesian deep learning.
\newblock \emph{CSUR}, 53\penalty0 (5):\penalty0 1--37, 2020.

\bibitem[Wang et~al.(2015)Wang, Wang, and Yeung]{CDL}
Wang, H., Wang, N., and Yeung, D.
\newblock Collaborative deep learning for recommender systems.
\newblock In \emph{KDD}, pp.\  1235--1244, 2015.

\bibitem[Wang et~al.(2019)Wang, Mao, He, Zhao, Jaakkola, and Katabi]{BIN}
Wang, H., Mao, C., He, H., Zhao, M., Jaakkola, T.~S., and Katabi, D.
\newblock Bidirectional inference networks: A class of deep bayesian networks
  for health profiling.
\newblock In \emph{AAAI}, volume~33, pp.\  766--773, 2019.

\bibitem[Wang et~al.(2020{\natexlab{a}})Wang, He, and Katabi]{CIDA}
Wang, H., He, H., and Katabi, D.
\newblock Continuously indexed domain adaptation.
\newblock In \emph{ICML}, 2020{\natexlab{a}}.

\bibitem[Wang et~al.(2020{\natexlab{b}})Wang, Menkovski, Wang, Du, and
  Pechenizkiy]{ICL}
Wang, Y., Menkovski, V., Wang, H., Du, X., and Pechenizkiy, M.
\newblock Causal discovery from incomplete data: A deep learning approach.
\newblock 2020{\natexlab{b}}.

\bibitem[Wheeler et~al.(2007)Wheeler, Barrett, Benson, Bryant, Canese,
  Chetvernin, Church, DiCuccio, Edgar, Federhen, et~al.]{bird_database}
Wheeler, D.~L., Barrett, T., Benson, D.~A., Bryant, S.~H., Canese, K.,
  Chetvernin, V., Church, D.~M., DiCuccio, M., Edgar, R., Federhen, S., et~al.
\newblock Database resources of the national center for biotechnology
  information.
\newblock \emph{Nucleic acids research}, 36\penalty0 (suppl\_1):\penalty0
  D13--D21, 2007.

\bibitem[Wu et~al.(2019)Wu, Winston, Kaushik, and Lipton]{wu2019domain}
Wu, Y., Winston, E., Kaushik, D., and Lipton, Z.
\newblock Domain adaptation with asymmetrically-relaxed distribution alignment.
\newblock In \emph{International conference on machine learning}, pp.\
  6872--6881. PMLR, 2019.

\bibitem[Xu et~al.(2022)Xu, He, Lee, Wang, and Wang]{GRDA}
Xu, Z., He, H., Lee, G.-H., Wang, Y., and Wang, H.
\newblock Graph-relational domain adaptation.
\newblock In \emph{ICLR}, 2022.

\bibitem[Xu et~al.(2023)Xu, Hao, He, and Wang]{VDI}
Xu, Z., Hao, G.-Y., He, H., and Wang, H.
\newblock Domain-indexing variational bayes: Interpretable domain index for
  domain adaptation.
\newblock In \emph{ICLR}, 2023.

\bibitem[Yang \& Yuen(2019)Yang and Yuen]{yang2019cross}
Yang, B. and Yuen, P.~C.
\newblock Cross-domain visual representations via unsupervised graph alignment.
\newblock In \emph{Proceedings of the AAAI Conference on Artificial
  Intelligence}, volume~33, pp.\  5613--5620, 2019.

\bibitem[Yang et~al.(2022)Yang, Yuan, Zhang, Wang, Chen, Liu, Tarolli, Crepeau,
  Bukartyk, Junna, Videnovic, Ellis, Lipford, Dorsey, and Katabi]{PDPredict}
Yang, Y., Yuan, Y., Zhang, G., Wang, H., Chen, Y.-C., Liu, Y., Tarolli, C.,
  Crepeau, D., Bukartyk, J., Junna, M., Videnovic, A., Ellis, T., Lipford, M.,
  Dorsey, R., and Katabi, D.
\newblock Artificial intelligence-enabled detection and assessment of
  parkinson’s disease using nocturnal breathing signals.
\newblock \emph{Nature medicine}, 1\penalty0 (1):\penalty0 1--1, 2022.

\bibitem[Zhang et~al.(2019)Zhang, Liu, Long, and Jordan]{MDD}
Zhang, Y., Liu, T., Long, M., and Jordan, M.~I.
\newblock Bridging theory and algorithm for domain adaptation.
\newblock \emph{arXiv preprint arXiv:1904.05801}, 2019.

\bibitem[Zhao et~al.(2018)Zhao, Zhang, Wu, Moura, Costeira, and Gordon]{AMSDA}
Zhao, H., Zhang, S., Wu, G., Moura, J. M.~F., Costeira, J.~P., and Gordon,
  G.~J.
\newblock Adversarial multiple source domain adaptation.
\newblock In \emph{NIPS}, pp.\  8568--8579, 2018.

\bibitem[Zhao et~al.(2019)Zhao, des Combes, Zhang, and Gordon]{InvariantDA}
Zhao, H., des Combes, R.~T., Zhang, K., and Gordon, G.~J.
\newblock On learning invariant representations for domain adaptation.
\newblock In \emph{ICML}, pp.\  7523--7532, 2019.

\bibitem[Zhao et~al.(2017)Zhao, Yue, Katabi, Jaakkola, and Bianchi]{CDANN}
Zhao, M., Yue, S., Katabi, D., Jaakkola, T.~S., and Bianchi, M.~T.
\newblock Learning sleep stages from radio signals: {A} conditional adversarial
  architecture.
\newblock In \emph{ICML}, pp.\  4100--4109, 2017.

\bibitem[Zhao et~al.(2020)Zhao, Hoti, Wang, Raghu, and Katabi]{MSA}
Zhao, M., Hoti, K., Wang, H., Raghu, A., and Katabi, D.
\newblock Assessment of medication self-administration using artificial
  intelligence.
\newblock \emph{Nature medicine}, 2020.

\bibitem[Zhuang et~al.(2020)Zhuang, Qi, Duan, Xi, Zhu, Zhu, Xiong, and
  He]{transfer_learning}
Zhuang, F., Qi, Z., Duan, K., Xi, D., Zhu, Y., Zhu, H., Xiong, H., and He, Q.
\newblock A comprehensive survey on transfer learning.
\newblock \emph{Proceedings of the IEEE}, 109\penalty0 (1):\penalty0 43--76,
  2020.

\bibitem[Zou et~al.(2018)Zou, Yu, Kumar, and Wang]{Zou_2018_ECCV}
Zou, Y., Yu, Z., Kumar, B.~V., and Wang, J.
\newblock Unsupervised domain adaptation for semantic segmentation via
  class-balanced self-training.
\newblock In \emph{Proceedings of the European Conference on Computer Vision
  (ECCV)}, September 2018.

\end{thebibliography}
